\documentclass{article}

\usepackage{fix-cm}
\usepackage{arxiv}

\usepackage[utf8]{inputenc}
\usepackage[T1]{fontenc}

\usepackage{hyperref}
\usepackage{url}
\usepackage{microtype}
\usepackage{nicefrac}

\usepackage{amsmath}
\usepackage{amsfonts}

\usepackage{graphicx}
\graphicspath{{./images/}{./Figs/}}

\usepackage[table]{xcolor}
\usepackage{booktabs}
\usepackage{array}
\usepackage{tabularx}
\usepackage{multirow}
\usepackage{adjustbox}

\usepackage{float}
\usepackage{placeins}
\usepackage{needspace}

\usepackage{ragged2e}

\usepackage{caption}

\newcommand{\smalltableformat}{%
\fontsize{10}{11}\selectfont
\setlength{\tabcolsep}{3.4pt}
\renewcommand{\arraystretch}{2}
}

\newcommand{\resulttableformat}{%
\fontsize{9}{10}\selectfont
\setlength{\tabcolsep}{2.8pt}
\renewcommand{\arraystretch}{1.5}
}
\title{ConceptSMILE: Auditing the Trustworthiness of Concept-Based Explainable AI}

\author{
Mohadeseh Mollapour \\
  School of Computer Science\\
  University of Hull\\
  Hull, United Kingdom \\
  \texttt{m.mollapour-papkyadeh-2023@hull.ac.uk} \\
   \And
 Koorosh Aslansefat\\
  School of Computer Science\\
  University of Hull\\
   Hull, United Kingdom\\
  \texttt{k.aslansefat@hull.ac.uk} \\
  \And
Zeinab Dehghani\\
PhD Researcher (Engineering), WMG\\
University of Warwick\\
Coventry, United Kingdom\\
\texttt{sara.dehghani@warwick.ac.uk}\\
  \And
 Bhupesh Kumar Mishra\\
  School of Computer Science \\
  University of Hull\\
 Hull, United Kingdom\\
  \texttt{bhupesh.mishra@hull.ac.uk} \\
\And
Tejal Shah\\
School of Computing\\
Newcastle University\\
Newcastle upon Tyne, United Kingdom\\
\texttt{tejal.shah@ncl.ac.uk}\\
\And
 Zhibao Mian\\
  School of Computer Science \\
  University of Hull\\
 Hull, United Kingdom\\
  \texttt{z.mian2@hull.ac.uk} \\
}

\begin{document}
\maketitle
\begin{abstract}
Concept-based explainable artificial intelligence (AI) can make model reasoning more human-understandable, but concept-level outputs are not automatically trustworthy. We introduce ConceptSMILE, a model-agnostic perturbation-based auditing framework for evaluating the reliability of concept-based explanations. Rather than replacing SMILE, ConceptSMILE extends its perturbation-based logic from feature or region level attribution to the auditing of human-understandable concept explanations. The framework perturbs input regions, measures concept-response shifts, applies locality weighting, and fits an XGBoost surrogate to approximate local concept behaviour. Reliability is assessed through attribution accuracy, surrogate fidelity, faithfulness, stability, and consistency. We evaluate ConceptSMILE on retinal fundus images by comparing MedSAM-derived visual concepts with VLM-based semantic concepts. Results show that reliability varies across concepts and pathways: MedSAM achieves stronger spatial attribution and the highest surrogate fidelity ($R^2 = 0.8503$, $R_w^2 = 0.8465$), while the VLM pathway shows stronger vessel faithfulness and stronger stability under selected artefact conditions. ConceptSMILE provides an independent audit layer for evaluating the trustworthiness of concept-based XAI.
\end{abstract}
\section{Introduction}
Artificial intelligence systems are increasingly used in high-stakes domains where model decisions must be not only accurate but also transparent, reliable, and accountable. Deep learning, vision transformers, multimodal models, and foundation models have achieved strong performance in areas such as medical imaging, autonomous systems, environmental monitoring, and decision support. However, these models often remain difficult to interpret because their internal reasoning is usually represented through high-dimensional latent features rather than human-understandable evidence~\cite{arrieta2020explainable}. This lack of transparency becomes particularly problematic when AI systems are used in settings where incorrect or poorly justified decisions may affect human safety, clinical judgement, or public trust~\cite{guidotti2018survey}.

Explainable Artificial Intelligence (XAI) has emerged to address this problem by making model behaviour more understandable to users and domain experts. Widely used post-hoc methods such as LIME and SHAP provide important tools for identifying influential input features~\cite{ribeiro2016should,lundberg2017unified}. In computer vision, Grad-CAM has also been widely used to highlight image regions that contribute to a model prediction~\cite{selvaraju2017grad}. These methods have been applied across many domains, including medical image analysis, where heatmaps and saliency maps can show which parts of an image contribute to a prediction. Nevertheless, region-based attribution is often insufficient for high-stakes decision-making. A heatmap may show where a model attends, but it does not necessarily explain what clinically, semantically, or operationally meaningful evidence the model has recognised. Prior work has also shown that attribution explanations can be fragile or fail basic reliability checks~\cite{adebayo2018sanity,ghorbani2019towards}. In high-stakes settings, this motivates the need for more interpretable and accountable explanation strategies~\cite{rudin2019stop}.

Concept-Based Explainable AI (C-XAI) has therefore become an important direction for improving interpretability. Instead of explaining predictions only through pixels, tokens, or low-level features, concept-based approaches aim to express model behaviour through human-understandable concepts~\cite{poeta2023concept}. These may include clinical biomarkers in medical imaging, objects and relations in visual reasoning, geographic units in spatial modelling, or semantic attributes in multimodal systems. Concept Activation Vectors introduced an influential way of testing how human-defined concepts affect model predictions~\cite{kim2018interpretability}. Concept Bottleneck Models later integrated concepts directly into the prediction pipeline, allowing models to reason through intermediate human-understandable variables~\cite{koh2020concept}. Prototype-based models provide another route by explaining predictions through similarity to learned representative examples~\cite{chen2019looks}. More recently, language-guided concept approaches have shown how textual concepts and vision-language alignment can support semantically richer explanations~\cite{yang2023language}. Together, these methods show that concepts can provide a more meaningful interface between complex models and human reasoning.

However, the presence of concepts does not automatically guarantee trustworthy explanations. A model may produce concept-level outputs that appear plausible while still relying on spurious correlations, unstable features, label leakage, concept entanglement, or incomplete representations~\cite{margeloiu2021concept}. Leakage in concept bottleneck models is particularly problematic because concept variables may encode hidden information about the target label rather than representing pure human-defined concepts~\cite{havasi2022addressing}. Other work has also highlighted the difficulty of obtaining robust and disentangled concept representations in practice~\cite{marconato2022glancenets}. In high-stakes domains, this creates a critical reliability gap. A concept explanation can be human-readable but still unfaithful to the model's actual behaviour. It may identify a clinically meaningful concept while its prediction is driven by an artefact, acquisition bias, or non-causal shortcut. Therefore, concept-based explanations should not be accepted as inherently self-explanatory. They require independent auditing.

To address this gap, we introduce ConceptSMILE (Concept-level Statistical Model-agnostic Interpretability with Local Explanations), a concept-level extension of SMILE designed to audit the trustworthiness of concept-based explanations. While SMILE focuses on local perturbation-based interpretability at the feature or region level, ConceptSMILE adapts this idea to human-understandable concepts. Specifically, it perturbs input regions, measures concept-response shifts, applies locality weighting, and fits a local surrogate model to evaluate whether concept-level explanations remain faithful, stable, and consistent under controlled perturbations~\cite{aslansefat2023smile}. It also follows the broader principle of local perturbation-based explanation, where changes around an input are used to understand model behaviour~\cite{ribeiro2016should}. Instead of asking only which input regions influence a model output, ConceptSMILE asks whether human-understandable concepts remain accurate, faithful, stable, and consistent when the input is systematically perturbed. The framework is model-agnostic and can be applied to any system that produces concept-level responses, including segmentation models, concept bottleneck models, vision-language models, multimodal systems, or other concept-based explanation pipelines.

\begin{figure}[H]
    \centering
    \includegraphics[width=0.9\linewidth]{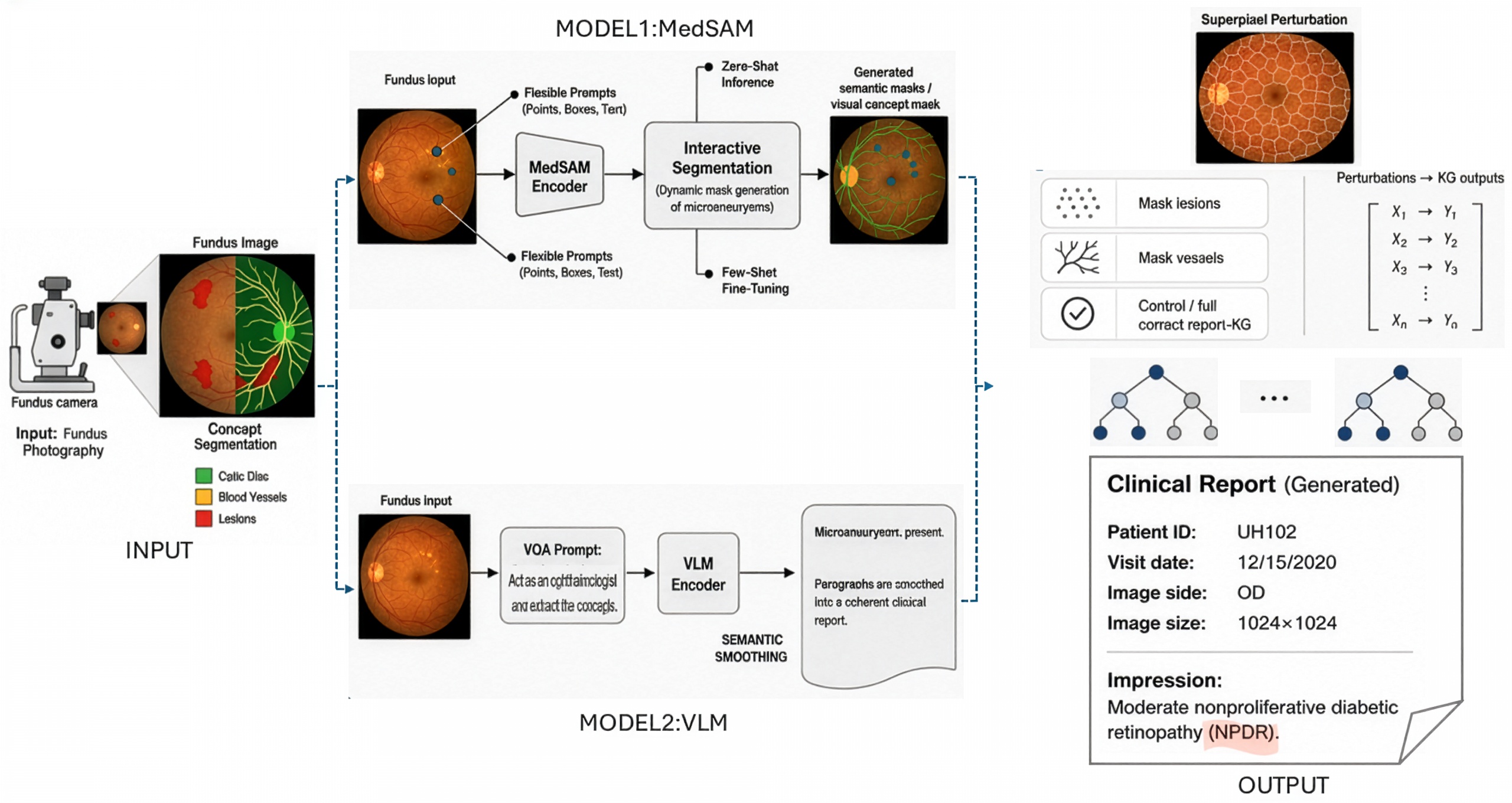}
    \caption[Integrated C-XAI pipeline for retinal analysis]{Integrated concept-based explainability pipeline for retinal analysis.}
    \label{fig:overview of model}
\end{figure}

ConceptSMILE operates by first extracting concept-level outputs from a target model or explanation pathway. It then generates controlled perturbations of the input, measures how each concept response changes under perturbation, assigns locality weights using distance functions such as cosine or Wasserstein distance, and fits a local surrogate model to approximate concept-response behaviour. In this study, an XGBoost surrogate is used to capture potentially non-linear relationships between perturbation patterns and concept-confidence shifts. The resulting audit allows the reliability of concept-level explanations to be evaluated through multiple dimensions: attribution accuracy, fidelity, faithfulness, stability, and consistency.
We demonstrate ConceptSMILE using retinal image analysis as a high-stakes medical case study. Retinal imaging is well suited for this validation because it contains clinically meaningful concepts such as lesions, blood vessels, and optic disc structures, and because clinical interpretation depends on whether AI systems use these concepts in a reliable way. However, retinal imaging is not the boundary of the method. It is used here as a proof-of-concept domain to show how ConceptSMILE can audit both visual and semantic concept explanations. Specifically, we compare a MedSAM-based pathway for segmentation-guided visual concepts with a vision-language-model pathway for semantic concept-level explanations. Fig.~\ref{fig:overview of model} summarises the integrated concept-based explainability pipeline used in this study, showing how retinal images are processed through concept pathways before ConceptSMILE auditing.

The key contributions of this paper are:
\begin{itemize}
\item \textbf{Concept-level auditing for C-XAI:} We introduce ConceptSMILE, a perturbation-based framework for evaluating the trustworthiness of concept-based explanations.
\item \textbf{Extension of SMILE to concept explanations:} We reformulate SMILE from superpixel-level attribution into concept-level reliability testing using controlled perturbations, locality weighting, and surrogate modelling.
\item \textbf{Model-agnostic validation through two pathways:} We demonstrate ConceptSMILE on retinal fundus images using two different concept-extraction pathways: MedSAM-based visual concepts and VLM-based semantic concepts, evaluated through attribution accuracy, fidelity, faithfulness, stability, and consistency.
\end{itemize}
\section{Literature Review}
This section reviews C-XAI and its role in making model reasoning more transparent, interpretable, and trustworthy ~\cite{arrieta2020explainable,guidotti2018survey,poeta2023concept}. We first summarise existing post-hoc and explainable-by-design C-XAI methods, then examine reliability challenges such as spurious correlation, concept leakage, instability, and weak faithfulness. These gaps motivate ConceptSMILE as a model-agnostic, perturbation-based auditing framework for concept-level explanations across domains, demonstrated through retinal image analysis as a high-stakes case study.
\subsection{Concept-Based Explainable AI}
C-XAI aims to explain model behaviour through human-understandable concepts rather than low-level features, pixels, or latent activations. This direction has emerged as a response to the limitations of conventional post-hoc attribution methods, which often identify influential regions but do not clearly describe what semantic evidence the model has used~\cite{arrieta2020explainable}. In contrast, concept-based methods seek to align machine reasoning with abstractions that are meaningful to humans, such as objects, attributes, clinical biomarkers, or domain-specific semantic units~\cite{poeta2023concept}.
Early concept-based approaches include Testing with Concept Activation Vectors (TCAV), which quantifies the influence of user-defined concepts on model predictions by measuring directional sensitivity in latent space~\cite{kim2018interpretability}. Automatic Concept-based Explanations (ACE) extended this idea by discovering concepts automatically from image segments, reducing the need for manually predefined concept sets~\cite{ghorbani2019towards}. Other post-hoc approaches analyse whether internal neurons, filters, or representation spaces correspond to interpretable concepts, providing insight into how trained models encode semantic information~\cite{bau2017network}.

A second family of methods integrates concepts directly into the model architecture. Concept Bottleneck Models (CBMs) force predictions to pass through an intermediate concept layer, making the decision process more transparent and potentially allowing human intervention at the concept level~\cite{koh2020concept}. More recent Concept Embedding Models (CEMs) relax the strict bottleneck structure by representing each concept through richer embeddings, aiming to improve the balance between interpretability and predictive performance~\cite{zarlenga2024does}. Prototype-based models provide another form of concept-level reasoning by explaining predictions through similarity to learned representative examples~\cite{chen2019looks}.
Despite these advances, concept-based explanations are not automatically trustworthy. A model may produce concepts that appear meaningful while still relying on spurious correlations, incomplete concept representations, or information leakage from the target label~\cite{margeloiu2021concept}. Post-hoc concept explanations may also depend strongly on the choice of probe dataset, concept examples, or representation layer, which can affect their stability and faithfulness~\cite{ramaswamy2023overlooked}. These limitations show that C-XAI should not be evaluated only by whether its concepts are human-readable, but also by whether those concepts remain reliable under systematic testing.

\begin{figure}[H]
    \centering
    \includegraphics[width=0.7\linewidth]{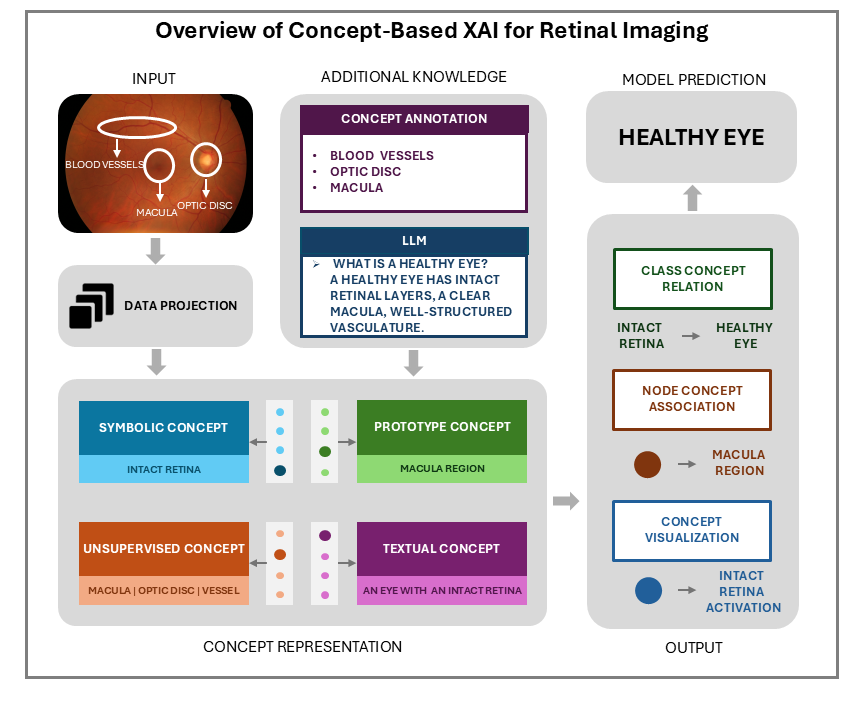}
    \caption{Overview of Concept-Based XAI for retinal imaging. The pipeline maps fundus images into human-understandable clinical concepts and uses them to support concept-level explanation and prediction.}
    \label{fig:concept based XAI}
\end{figure}

This reliability gap motivates ConceptSMILE. Rather than treating concept-based explanations as inherently self-explanatory, ConceptSMILE evaluates how concept-level outputs behave under controlled perturbations. In this way, it extends concept-based XAI from semantic interpretability toward measurable trustworthiness. Fig.~\ref{fig:concept based XAI} illustrates how retinal fundus images are mapped into human-understandable clinical concepts to support concept-level explanation and prediction.
\subsubsection{Challenges in Concept-Based XAI}
\begin{figure}[H]
    \centering
    \includegraphics[width=0.9\linewidth]{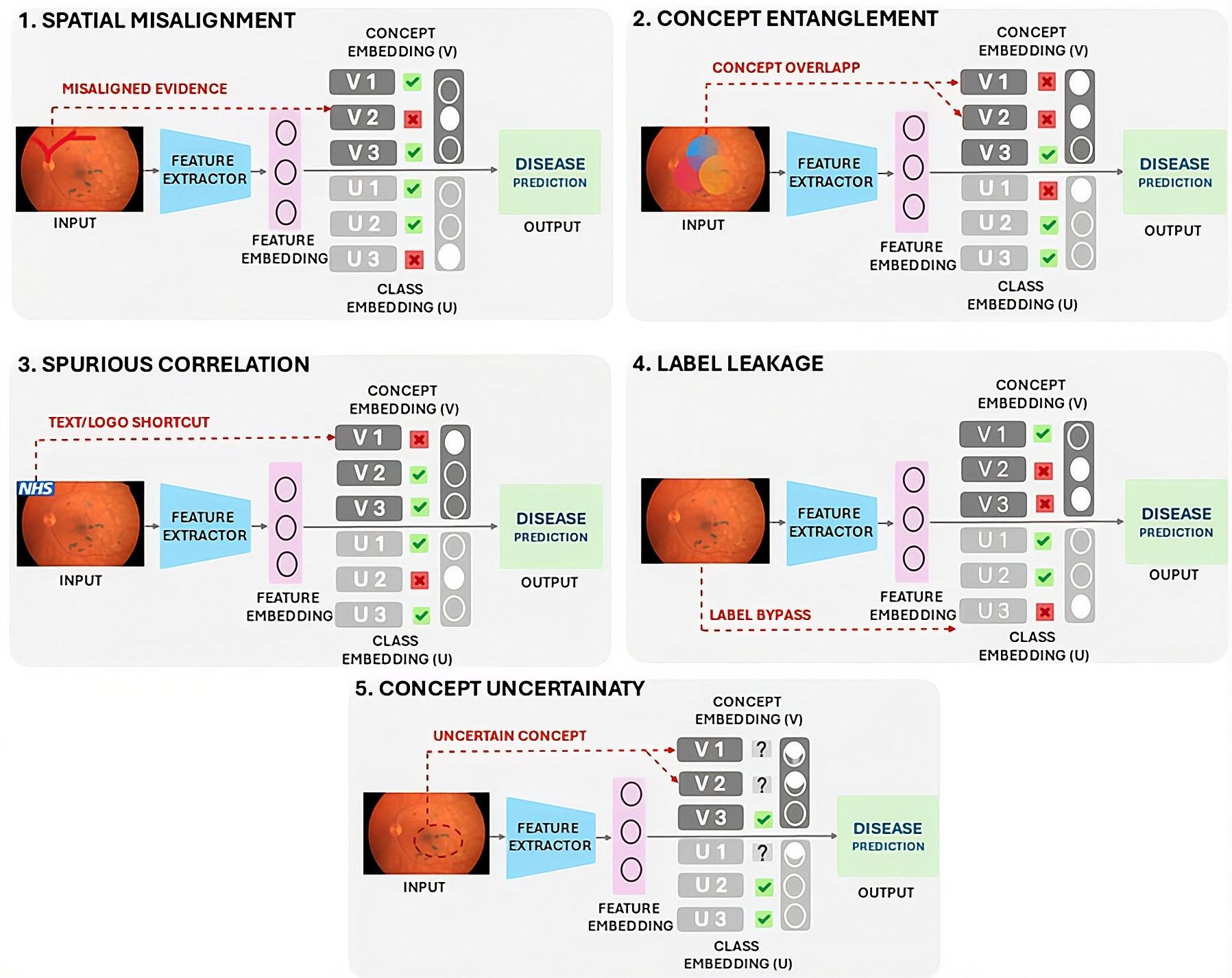}
    \caption{Key reliability challenges in Concept-Based XAI, including spatial misalignment, concept entanglement, spurious correlation, label leakage, and concept uncertainty.}
    \label{fig:gaps}
\end{figure}

Although Concept-Based Explainable AI provides a more human-understandable form of interpretation, concept-level explanations should not be assumed to be inherently reliable. A model may produce concepts that appear meaningful to humans while still relying on incomplete, unstable, or non-causal evidence. This is particularly important in high-stakes domains, where an explanation must be more than plausible; it must be faithful to the model's actual behaviour and robust under changes to the input~\cite{rudin2019stop}. Fig.~\ref{fig:gaps} summarises the main reliability challenges considered in this work, including spatial misalignment, concept entanglement, spurious correlation, label leakage, and concept uncertainty.

One major challenge is spurious correlation. In this case, a model may associate a concept or prediction with an irrelevant shortcut that happens to co-occur with the target class. In image-based domains, such shortcuts may include acquisition artefacts, borders, text marks, logos, camera-specific patterns, or other dataset-dependent cues~\cite{brown2023making}. As a result, a concept explanation may appear semantically correct while the underlying model behaviour is driven by non-relevant evidence.

A second challenge is concept leakage, where the concept layer does not operate as a clean explanatory bottleneck. Instead of representing only the intended human-defined concept, concept variables may encode hidden information about the final target label~\cite{mahinpei2021promises}. This can make an explanation appear clinically or semantically meaningful even when the model is partly using information that is not contained in the stated concept. Such leakage weakens the faithfulness of concept-based explanations and makes their interpretation less reliable~\cite{margeloiu2021concept}.

A further challenge is instability. Previous work has shown that explanations can change under small perturbations, changes in probe data, or adversarial manipulation~\cite{ghorbani2019interpretation}. Concept-based explanations can also depend on the selected concept examples, representation layer, or probing dataset, which may lead to different explanations for the same model~\cite{ramaswamy2023overlooked}. This instability is problematic because a trustworthy explanation should remain consistent when the underlying evidence has not meaningfully changed.

Concept uncertainty is another important limitation. Many concept-based methods treat concepts as fixed binary labels or deterministic scores, even when the visual or semantic evidence is ambiguous. In real-world settings, concepts may be overlapping, partially visible, or difficult to separate from related features. Probabilistic concept-based models attempt to address this issue by representing uncertainty at the concept level, but uncertainty-aware evaluation remains an open challenge~\cite{kim2023probabilistic}.

Together, these limitations show that human-understandable concepts do not automatically guarantee trustworthy explanations. Concept-based XAI therefore requires independent reliability testing that can examine whether concepts remain accurate, faithful, stable, and consistent under controlled perturbations. This motivates ConceptSMILE as a model-agnostic auditing framework for evaluating concept-level explanations rather than accepting them at face value.
\subsubsection{Research Gap: From Self-Explanation to Concept-Level Auditing}
Concept-based models are often presented as more interpretable than conventional black-box models because they expose intermediate concepts that can be inspected by humans. Explainable-by-design approaches, such as Concept Bottleneck Models, make this idea explicit by forcing predictions to pass through a concept layer before reaching the final output~\cite{koh2020concept}. However, the presence of a concept layer does not mean that the model is automatically self-explanatory, self-verifying, or clinically trustworthy. A model can provide a human-readable explanation while still relying on hidden shortcuts, incomplete concept representations, or non-conceptual information.

This limitation is particularly important for self-explainable and concept-bottleneck architectures. Although these models appear transparent, previous studies have shown that concept representations may leak information about the final class label, weakening the assumption that the concept layer is a pure explanatory bottleneck~\cite{margeloiu2021concept}. Similarly, concept variables may encode unintended information that improves prediction performance but reduces the faithfulness of the explanation~\cite{havasi2022addressing}. Therefore, even models that are designed to be interpretable still require independent reliability assessment.

Post-hoc concept-based methods face a related problem. Methods such as TCAV and ACE can reveal whether a trained model is sensitive to human-defined or automatically discovered concepts~\cite{kim2018interpretability}. However, these explanations may depend on the choice of concept examples, probe dataset, representation layer, or perturbation setting~\cite{ramaswamy2023overlooked}. This means that concept-level explanations may change even when the underlying model remains the same, raising concerns about stability and reproducibility.

Existing evaluation practices remain fragmented. Some metrics assess concept accuracy, others examine completeness, faithfulness, robustness, or human interpretability, but these are often applied separately and do not provide a unified audit of concept-level reliability~\cite{yeh2020completeness}. As a result, there remains a gap between producing concept-based explanations and verifying whether those explanations are accurate, faithful, stable, and consistent under controlled testing.

This gap motivates ConceptSMILE. Rather than assuming that concept-based explanations are reliable because they are human-readable, ConceptSMILE treats them as claims that must be audited. It provides a model-agnostic framework for testing how concept-level explanations respond to controlled perturbations, allowing both visual and semantic concept pathways to be evaluated under the same reliability protocol.

\subsection{Explainable AI and SMILE/LIME-Based Methods}
Post-hoc explainable AI methods aim to explain the behaviour of already trained models without modifying their internal architecture. Widely used approaches such as LIME, SHAP, and Grad-CAM provide different mechanisms for identifying influential input features or regions~\cite{ribeiro2016should,lundberg2017unified,selvaraju2017grad}. LIME explains individual predictions by generating local perturbations around an input and fitting a simple surrogate model to approximate the behaviour of the original model in that local neighbourhood~\cite{ribeiro2016should}. SHAP estimates feature contributions using a game-theoretic formulation~\cite{lundberg2017unified}, while Grad-CAM produces class-discriminative visual heatmaps by using gradients from convolutional feature maps~\cite{selvaraju2017grad}.

These methods have been valuable for improving the transparency of black-box models, but they mainly explain low-level features, pixels, regions, or tokens. In high-stakes settings, this is often insufficient because an explanation may indicate where the model attends without showing whether the model uses meaningful concepts in a reliable way~\cite{adebayo2018sanity,rudin2019stop}. This limitation is particularly important for concept-based XAI, where the explanation is expected to represent human-understandable evidence rather than only local feature importance~\cite{poeta2023concept}.

Recent extensions of LIME have attempted to improve local explanation quality through more specialised perturbation strategies, sampling mechanisms, surrogate models, and domain-specific adaptations. Some variants focus on improving the sampling strategy and perturbation process. For example, Anchor explanations improve rule-based local precision~\cite{ribeiro2018anchors}, S-LIME improves explanation stability~\cite{zhou2021s}, US-LIME uses uncertainty sampling to increase fidelity~\cite{saadatfar2024us}, Guided-LIME introduces structured sampling~\cite{sangroya2020guided}, and DLIME provides deterministic local explanations for improved stability~\cite{zafar2021deterministic}. Other work has examined the role of locality definition in post-hoc surrogate explanations~\cite{laugel2018defining}. 

A second group of methods modifies the local surrogate or weighting mechanism used by LIME. QLIME introduces a quadratic local surrogate~\cite{bramhall2020qlime}, BayLIME applies Bayesian local explanation~\cite{zhao2021baylime}, and ALIME uses an autoencoder-based approach for local interpretability~\cite{shankaranarayana2019alime}. A further group focuses on stability, fidelity, optimisation, or computational efficiency, including OptiLIME~\cite{visani2020optilime} and GLIME~\cite{tan2023glime}. In addition, LIME-style methods have been adapted to different data modalities, including graph data through GraphLIME~\cite{huang2022graphlime}, time-series forecasting through TS-MULE~\cite{schlegel2021ts}, audio/music analysis~\cite{mishra2017local}, and ECG classification~\cite{abdullah2023b}. Recent surveys provide broader taxonomies of these variants and their technical differences~\cite{knab2025lime}. 

For image-based explanation, segmentation-aware variants such as DSEG-LIME aim to generate more meaningful perturbation regions by improving the relationship between image segmentation and local explanation quality~\cite{knab2024dseg}. Other extensions, such as SLICE, focus on improving the stability and consistency of local image explanations~\cite{bora2024slice}. More broadly, recent reviews of LIME variants show that perturbation design, locality definition, and surrogate-model selection can strongly affect explanation reliability~\cite{knab2025whichlime}. These developments are relevant to ConceptSMILE because concept-level auditing also depends on how local perturbations are generated, weighted, and approximated by a surrogate model.

SMILE extends the logic of local model-agnostic explanation by using perturbation-based sampling, distance-based locality weighting, and surrogate modelling to estimate how input regions influence model behaviour~\cite{aslansefat2023smile}. SMILE-style explanation has also been adapted to generative and multimodal settings. For instruction-based image editing, Image Editing SMILE applies prompt-side perturbations to examine how individual instruction tokens influence the generated or edited image, supporting word-level explanation for instruction-based image-editing models~\cite{dehghani2024mapping}. More recently, KG-SMILE extended this idea to knowledge-graph retrieval-augmented generation by perturbing graph entities and relations, measuring semantic response changes, and fitting local surrogate explanations to identify influential graph components~\cite{moghaddam2025explainable}. gSMILE further extended this idea to generative language models by perturbing input prompts, measuring semantic output shifts, using Wasserstein-based distance measures, and fitting local surrogate explanations~\cite{dehghani2025gsmile}. Related work has also explored black-box LLM interpretation through concept-level surrogate modelling, where sentence-level units are treated as interpretable concepts and an energy-based surrogate is used to estimate prompt response attribution under closed-model constraints~\cite{asgarinezhadinterpreting}. Wasserstein-based distances are grounded in optimal transport theory and provide a principled way to compare distributional shifts between original and perturbed representations~\cite{peyre2019computational}. These developments show that SMILE-style explanation can be adapted beyond conventional image classification to more complex model behaviours.

However, existing LIME, SMILE, Image Editing SMILE, KG-SMILE, and gSMILE-style methods primarily focus on input-level, output-level, token-level, prompt-level, or graph-component-level attribution. Although these methods improve local explanation across different modalities, they do not directly audit whether human-understandable concepts remain reliable under controlled perturbation. This motivates ConceptSMILE, which reformulates the SMILE-style perturbation framework from feature-, token-, prompt-, or component-level attribution into concept-level reliability auditing. Instead of asking only which pixels, regions, tokens, prompts, or graph components influence an output, ConceptSMILE asks whether visual or semantic concepts remain accurate, faithful, stable, and consistent when the input evidence is systematically changed.

\subsection{Explainable AI for Retinal and Medical Foundation Models}

Retinal image analysis is a clinically important setting for explainable AI because diagnostic reasoning depends on visible anatomical and pathological concepts. Fundus images contain clinically meaningful evidence such as lesions, blood vessels, the optic disc, macular regions, haemorrhages, exudates, and other biomarkers. Deep learning systems have shown strong performance for detecting diabetic retinopathy and related retinal disease patterns from fundus photographs~\cite{gulshan2016development,abramoff2018pivotal}. However, retinal explanations should not only highlight image regions, but should also indicate whether the model is using clinically meaningful concepts in a reliable way.

Recent medical foundation models and segmentation models provide new opportunities for concept-level explanation. RETFound demonstrates that foundation models can learn generalisable retinal representations from large-scale unlabelled retinal images~\cite{zhou2023retfound}. MedSAM enables segmentation-guided extraction of medically meaningful visual regions and can be adapted to identify retinal structures and abnormal regions. In a concept-based explanation setting, such segmentation outputs can be interpreted as visual concepts with spatial location, confidence, and graph-based structure.

Vision--language models provide a complementary form of explanation by generating semantic responses from retinal images using structured prompts. Ophthalmology-specific vision--language foundation models, such as VisionUnite, show the growing role of language-guided multimodal reasoning in ophthalmic diagnosis and explanation~\cite{li2025visionunite}. In this paper, the VLM pathway uses Qwen2.5-VL to extract semantic concept responses for predefined retinal concepts such as lesions, blood vessels, and the optic disc. This allows the same retinal image to be represented through both segmentation-guided visual evidence and language-guided semantic evidence.

Despite these opportunities, medical foundation models and VLMs can produce explanations that appear clinically plausible without being reliable. A generated concept may be unstable under perturbation, sensitive to irrelevant artefacts, weakly aligned with image evidence, or inconsistent across repeated runs. Similarly, segmentation-based visual concepts may be spatially grounded but still fail to show faithful perturbation--response behaviour. Therefore, visual and semantic concept explanations require independent auditing rather than being accepted as self-explanatory~\cite{adebayo2018sanity,ghorbani2019interpretation}.

This gap motivates ConceptSMILE as a unified reliability-auditing framework for retinal concept explanations. By applying the same perturbation, locality weighting, surrogate modelling, and evaluation protocol to both MedSAM-based visual concepts and VLM-based semantic concepts, ConceptSMILE provides a systematic way to compare explanation reliability across different concept-extraction pathways.
\section{Problem Definition}

Concept-based explanation methods aim to express model behaviour through human-understandable concepts rather than raw pixels, low-level features, or latent activations. However, a concept-level explanation is not necessarily reliable simply because it is interpretable. A model may report clinically or semantically meaningful concepts while its behaviour is still influenced by hidden shortcuts, unstable evidence, incomplete concept representations, or non-conceptual information~\cite{margeloiu2021concept,havasi2022addressing}. Therefore, the central problem addressed in this work is how to independently evaluate whether a concept-based explanation is accurate, faithful, stable, and consistent under controlled perturbations.

Let $f$ denote a fixed predictive model or explanation pathway, and let $x$ be an input sample. A concept-based explanation pathway produces a set of concept responses, denoted by
\begin{equation}
\mathcal{C}(x)=\{c_1(x),c_2(x),\ldots,c_m(x)\},
\label{eq:concept_response_set}
\end{equation}
where each $c_i(x)$ represents the presence, confidence, or importance of a human-understandable concept. In this work, such concepts may be visual concepts extracted from segmentation-based models or semantic concepts extracted from vision-language models. The objective is not to train a new concept model, but to audit whether the concepts produced by an existing pathway behave reliably when the input evidence is systematically changed. The concept-response set in Eq.~\eqref{eq:concept_response_set} provides the basis for this audit.

To perform this audit, a set of perturbed samples $\{x^{(k)}\}_{k=1}^{K}$ is generated from the original input $x$. Each perturbation modifies selected input regions while preserving the overall structure of the sample. The response shift of concept $c_i$ under perturbation $k$ is defined as:
\begin{equation}
\Delta c_i^{(k)} = c_i(x) - c_i\left(x^{(k)}\right).
\label{eq:concept_response_shift}
\end{equation}

As shown in Eq.~\eqref{eq:concept_response_shift}, the response shift measures how much the concept response changes after perturbation. If a concept is genuinely connected to the model's behaviour, perturbing evidence relevant to that concept should produce a measurable and consistent response shift. Conversely, if the concept response remains unchanged, changes unpredictably, or reacts strongly to irrelevant regions, the explanation may be unreliable.

The problem can therefore be formulated as concept-level reliability auditing: given an input $x$, a fixed model or explanation pathway $f$, and a set of concept responses $\mathcal{C}(x)$ defined in Eq.~\eqref{eq:concept_response_set}, determine whether the resulting concept-level explanation remains locally faithful, reproducible, and robust under controlled perturbation. ConceptSMILE addresses this problem by combining perturbation-based sampling, locality weighting, surrogate modelling, and multi-dimensional reliability evaluation.

\section{ConceptSMILE Framework}
\begin{figure}[H]
    \centering
    \includegraphics[width=0.8\linewidth]{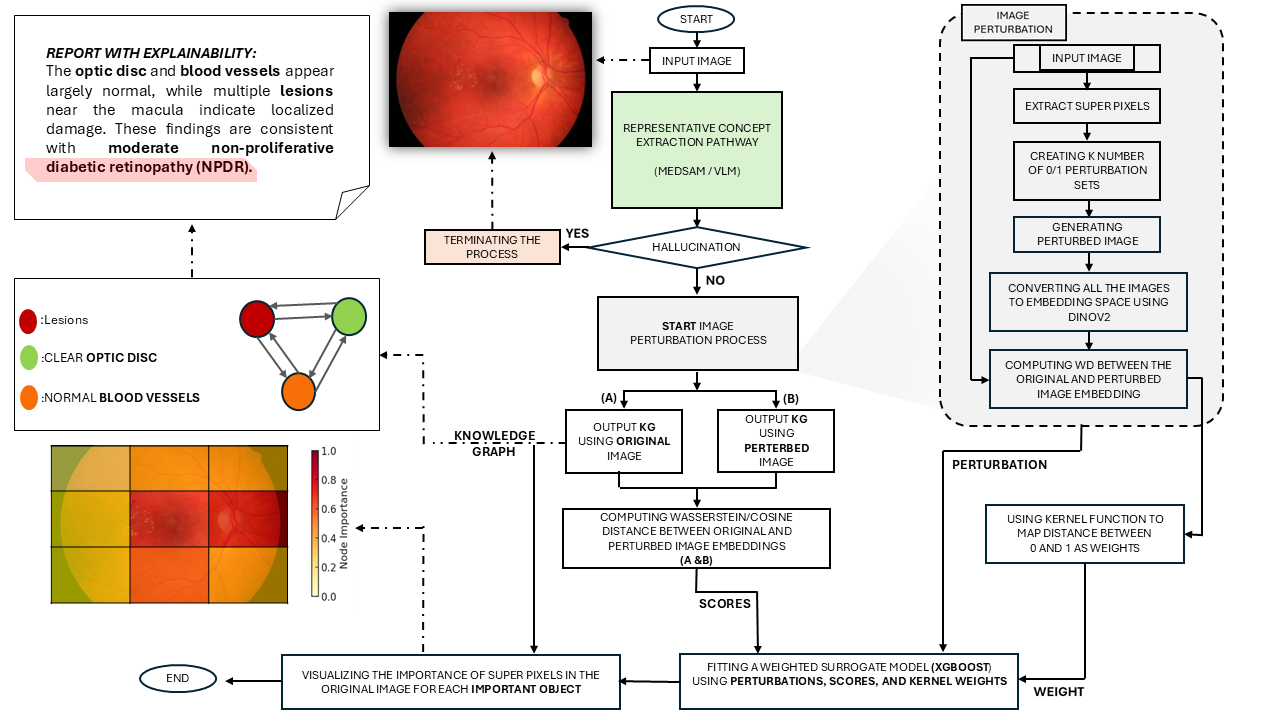}
    \caption{Overview of the ConceptSMILE framework. The pipeline extracts concept-level outputs, applies superpixel perturbations, computes locality weights, and fits a surrogate model for reliability auditing.}
    \label{fig:conceptsmile_overview}
\end{figure}

This section presents the proposed ConceptSMILE framework for auditing the reliability of concept-based explanations. Building on the problem definition introduced above, ConceptSMILE is designed as a model-agnostic pipeline that evaluates how concept-level explanations behave under controlled perturbations. As illustrated in Fig.~\ref{fig:conceptsmile_overview}, the framework first extracts human-understandable concepts from an existing explanation pathway, then generates perturbed versions of the input, measures concept-response shifts, applies locality weighting, and fits a surrogate model to approximate local concept behaviour. 
The resulting explanations are then assessed using multiple reliability dimensions, including attribution accuracy, fidelity, faithfulness, stability, and consistency. In this study, ConceptSMILE is demonstrated through two different concept-extraction pathways: MedSAM-based visual concepts and vision-language-model-based semantic concepts.
\subsection{Concept Extraction Pathways}

ConceptSMILE can be applied to explanation pathways that produce concept-level outputs. In this study, two complementary pathways are used to demonstrate this model-agnostic design: a MedSAM-based visual pathway and a vision-language-model-based semantic pathway. These pathways are not treated as competing diagnostic classifiers, but as two different mechanisms for generating concept-level explanations that can be audited under the same ConceptSMILE protocol.

In the visual pathway, MedSAM is used to extract segmentation-guided retinal concepts from fundus images~\cite{ma2024segment}. The extracted concepts correspond to clinically meaningful visual evidence, including lesion regions, blood vessels, and the optic disc. Each concept is represented through visual outputs such as segmentation masks, spatial regions, and concept-response scores. This pathway allows ConceptSMILE to examine whether visual concept explanations change in a meaningful way when the corresponding retinal evidence is perturbed.

In the semantic pathway, a vision-language model is used to generate concept-level responses from the same retinal images~\cite{bai2025qwen3}. A structured prompt is used to ask the model about the presence and relevance of predefined retinal concepts, including lesions, blood vessels, and optic disc structures. The same prompt structure and inference settings are kept fixed for the original and perturbed images so that any observed change in the semantic response can be attributed to the visual perturbation rather than to prompt variation.

Although the two pathways represent retinal evidence differently, both are analysed using the same ConceptSMILE pipeline. For each image, concept responses are first extracted from the original input and then re-computed after controlled perturbations. The resulting response changes provide the basis for measuring whether each concept-level explanation is accurate, faithful, stable, and consistent under systematic testing.

\subsection{Perturbation-Based Concept Auditing}

\begin{figure}[H]
    \centering
    \includegraphics[width=0.8\linewidth]{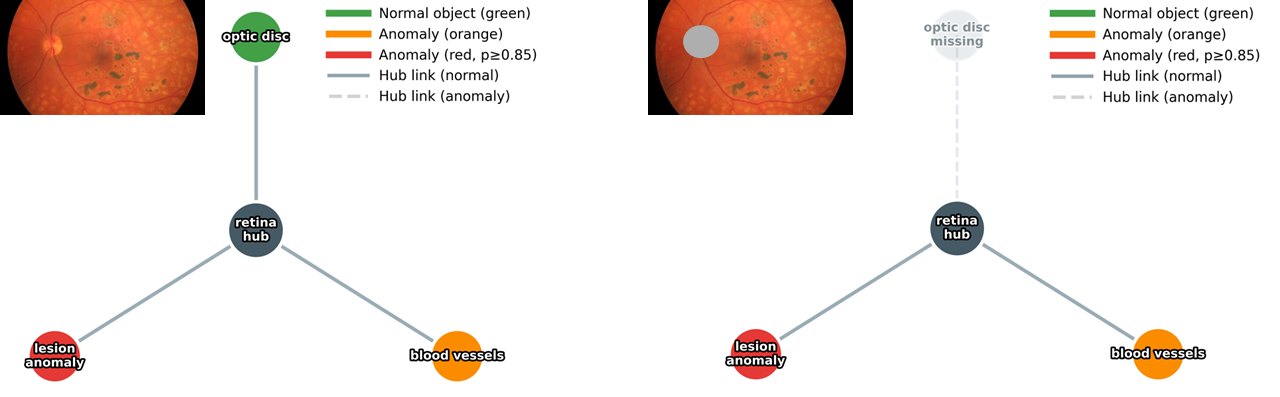}
    \caption{Perturbation-based concept graph auditing in ConceptSMILE. Masking the optic-disc region changes the corresponding concept node and illustrates how concept confidence responds to evidence removal.}
    \label{fig:concept_graph_auditing}
\end{figure}

After concept responses are extracted from the original input, ConceptSMILE applies controlled perturbations to examine whether each concept-level explanation is genuinely linked to the visual evidence in the image. The purpose of this stage is not to degrade the image arbitrarily, but to create a local neighbourhood around the original sample and observe how concept responses change when selected image regions are preserved or removed. The perturbation-based auditing process is illustrated in Fig.~\ref{fig:concept_graph_auditing}.

Let the original input image be denoted as $I^{(0)}$. The image is first divided into $M$ non-overlapping superpixel regions:
\begin{equation}
\mathcal{S}=\{S_1,S_2,\ldots,S_M\},
\label{eq:superpixel_set}
\end{equation}
where each $S_m$ represents one local image region. The superpixel set in Eq.~\eqref{eq:superpixel_set} defines the local regions used for perturbation.

A perturbation is represented by a binary vector:
\begin{equation}
z^{(k)} \in \{0,1\}^{M},
\label{eq:binary_perturbation_vector}
\end{equation}
where $z_m^{(k)}=1$ indicates that the $m$-th superpixel is preserved and $z_m^{(k)}=0$ indicates that it is masked. Thus, Eq.~\eqref{eq:binary_perturbation_vector} specifies which image regions are retained or removed in the $k$-th perturbation.

Applying the perturbation vector $z^{(k)}$ to the original image generates the $k$-th perturbed image $I^{(k)}$. Repeating this process produces the perturbed image set:
\begin{equation}
\mathcal{I}_{\mathrm{pert}}=\{I^{(1)},I^{(2)},\ldots,I^{(K)}\}.
\label{eq:perturbed_image_set}
\end{equation}
The set in Eq.~\eqref{eq:perturbed_image_set} represents the local neighbourhood of the original image used for concept-level auditing.

Each perturbed image is then passed through the same concept-extraction pathway used for the original image. The model pathway, prompt structure, and inference settings are kept fixed so that any observed change in the concept response can be attributed to the perturbation rather than to changes in the model configuration.

For a concept $c_i$, the response of the model pathway to the original image is denoted as $y_i^{(0)}$, while the response to the $k$-th perturbed image is denoted as $y_i^{(k)}$. The perturbation-induced concept-response shift is defined as:
\begin{equation}
\Delta y_i^{(k)} = y_i^{(0)} - y_i^{(k)}.
\label{eq:concept_shift}
\end{equation}

As shown in Eq.~\eqref{eq:concept_shift}, the magnitude of $\Delta y_i^{(k)}$ indicates how strongly the concept response changes when specific image regions are removed. If masking regions relevant to a concept produces a clear response shift, this suggests that the concept explanation is locally meaningful. In contrast, if the concept changes strongly when irrelevant or non-clinical regions are perturbed, the explanation may be unstable or affected by shortcut behaviour.

This perturbation-based audit allows ConceptSMILE to treat concept explanations as testable claims rather than fixed outputs. Instead of assuming that a concept is reliable because it is human-understandable, the framework evaluates whether the concept behaves consistently with the visual evidence under controlled perturbation.

\subsection{Locality Weighting and Surrogate Modelling}

After generating perturbed samples and measuring concept-response shifts, ConceptSMILE assigns a locality weight to each perturbation. The purpose of this step is to preserve the local nature of the explanation by giving higher importance to perturbed images that remain close to the original input and lower importance to heavily modified samples.

The original image and each perturbed image are first projected into an embedding space using an image representation function $\phi(\cdot)$:
\begin{equation}
e^{(0)}=\phi\left(I^{(0)}\right),
\qquad
e^{(k)}=\phi\left(I^{(k)}\right).
\label{eq:image_embeddings}
\end{equation}

As shown in Eq.~\eqref{eq:image_embeddings}, $e^{(0)}$ denotes the embedding of the original image, while $e^{(k)}$ denotes the embedding of the $k$-th perturbed image. These embeddings provide the representation space used to measure the locality of each perturbation.

The distance between the original and perturbed image embeddings is then computed as:
\begin{equation}
d^{(k)} = D\left(e^{(0)},e^{(k)}\right),
\label{eq:embedding_distance}
\end{equation}
where $D(\cdot,\cdot)$ denotes the selected distance function. In this work, cosine distance and Wasserstein distance are used to measure how far each perturbed image is from the original input in the embedding space. The distance definition in Eq.~\eqref{eq:embedding_distance} is used to quantify the locality of each perturbation.

The distance value is converted into a locality weight using an exponential locality kernel:
\begin{equation}
w^{(k)}=\exp\left(-\frac{\left(d^{(k)}\right)^2}{\sigma^2}\right),
\label{eq:locality_weight}
\end{equation}
where $\sigma$ is the kernel width parameter. As shown in Eq.~\eqref{eq:locality_weight}, perturbations that remain visually or semantically close to the original image receive higher weights, while distant perturbations contribute less to the local explanation.

For each concept $c_i$, ConceptSMILE fits a local XGBoost surrogate model to approximate the relationship between perturbation patterns and concept-response shifts. The surrogate receives the perturbation vector $z^{(k)}$ defined in Eq.~\eqref{eq:binary_perturbation_vector} as input and predicts the corresponding concept-response shift $\Delta y_i^{(k)}$ defined in Eq.~\eqref{eq:concept_shift}. The weighted optimisation objective is defined as:
\begin{equation}
g_i^{\mathrm{XGB}}
=
\arg\min_{g \in \mathcal{G}_{\mathrm{XGB}}}
\sum_{k=1}^{K}
w^{(k)}
\left[
\Delta y_i^{(k)} - g\left(z^{(k)}\right)
\right]^2 .
\label{eq:xgboost_surrogate}
\end{equation}

Here, $g_i^{\mathrm{XGB}}$ denotes the local surrogate fitted for concept $c_i$, and $\mathcal{G}_{\mathrm{XGB}}$ denotes the class of XGBoost regression models. The objective in Eq.~\eqref{eq:xgboost_surrogate} gives higher importance to perturbations that are closer to the original image through the locality weights $w^{(k)}$ defined in Eq.~\eqref{eq:locality_weight}. The fitted surrogate estimates which image regions most strongly influence each concept response under perturbation. In this way, ConceptSMILE converts perturbation behaviour into a local, concept-level explanation that can be quantitatively evaluated.

\subsection{Reliability Evaluation Metrics}

\begin{figure}[H]
    \centering
    \includegraphics[width=0.8\linewidth]{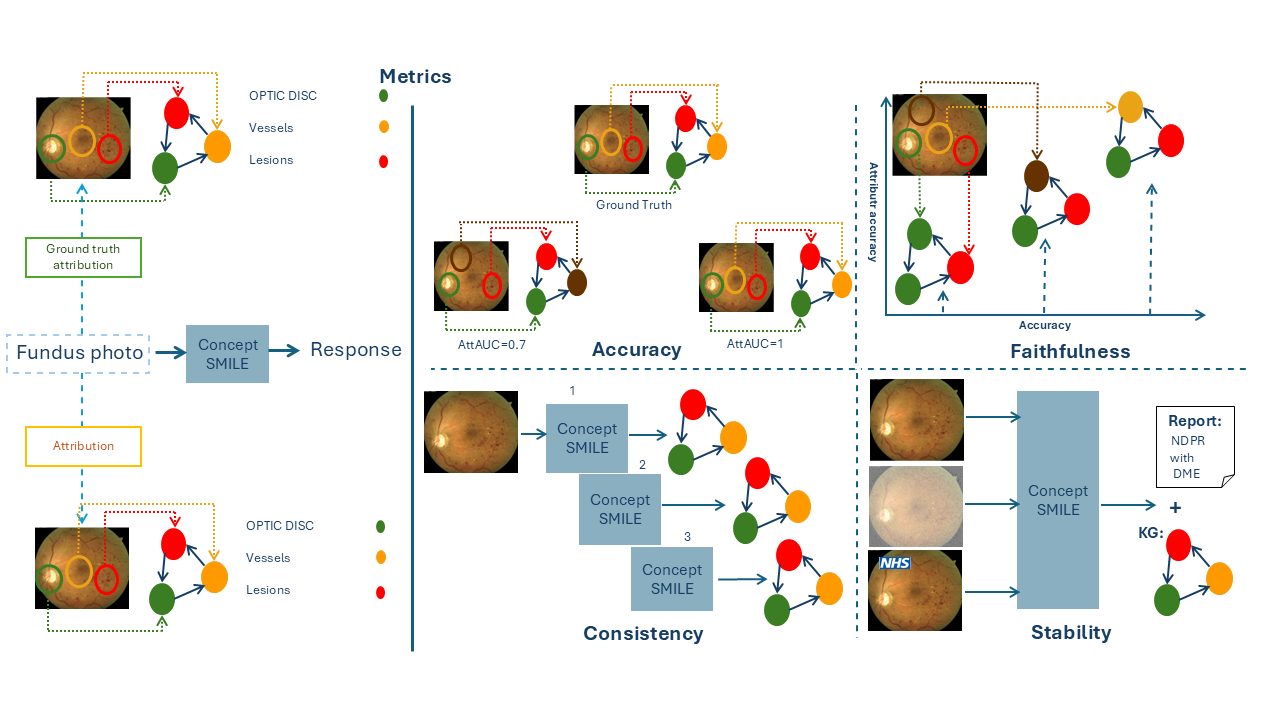}
    \caption{ConceptSMILE evaluation framework for auditing concept-level explanations through attribution accuracy, fidelity, faithfulness, stability, and consistency.}
    \label{fig:evaluation}
\end{figure}

After fitting the local surrogate model, ConceptSMILE evaluates whether the resulting concept-level explanations are reliable. The evaluation is designed to move beyond visual or semantic plausibility and instead test whether the extracted concepts are accurate, locally faithful, reproducible, and robust under controlled conditions. In this work, five complementary reliability dimensions are used: attribution accuracy, fidelity, faithfulness, stability, and consistency. The evaluation framework is illustrated in Fig.~\ref{fig:evaluation}.

\textbf{Attribution accuracy} measures whether the regions or concepts identified as important by ConceptSMILE correspond to clinically meaningful evidence. Given an attribution score $\hat{a}_j$ for concept element $j$ and a binary reference label $y_j \in \{0,1\}$, the predicted attribution label is obtained using a threshold $\tau$:
\begin{equation}
\hat{y}_j =
\begin{cases}
1, & \text{if } \hat{a}_j \geq \tau, \\
0, & \text{if } \hat{a}_j < \tau.
\end{cases}
\label{eq:binary_attribution}
\end{equation}

As shown in Eq.~\eqref{eq:binary_attribution}, attribution scores are converted into binary attribution labels using the threshold $\tau$. Attribution accuracy is then evaluated using ATT ACC, ATT F1, and ATT AUROC. These metrics assess whether clinically relevant concept regions receive higher attribution scores than irrelevant or weakly related regions.

\textbf{Fidelity} evaluates how well the local surrogate model approximates the concept-response shifts observed under perturbation. A high-fidelity surrogate should closely predict the observed shift $\Delta y_i^{(k)}$ defined in Eq.~\eqref{eq:concept_shift} from the perturbation vector $z^{(k)}$ defined in Eq.~\eqref{eq:binary_perturbation_vector}. In this work, fidelity is measured using error-based and coefficient-based metrics, including MSE, MAE, weighted MSE, weighted MAE, $R^2$, and weighted $R^2$. Higher $R^2$ values and lower error values indicate stronger local surrogate fidelity.

\begin{figure}[H]
    \centering
    \includegraphics[width=0.95\linewidth]{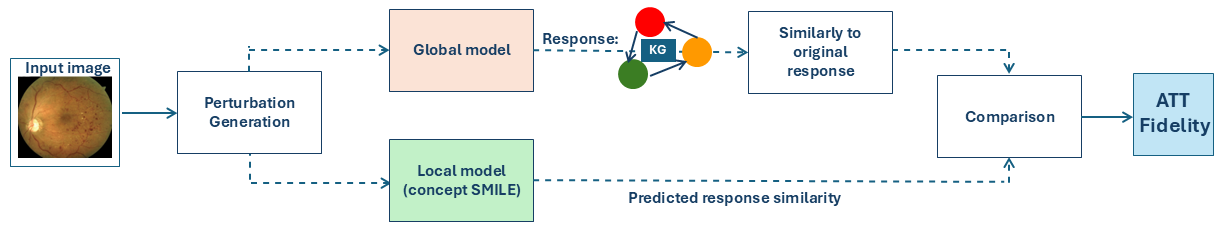}
    \caption{Fidelity evaluation pipeline. The observed concept-response shifts from the original pathway are compared with the surrogate-predicted shifts to assess local explanation fidelity.}
    \label{fig:fidelity}
\end{figure}

Fig.~\ref{fig:fidelity} illustrates the fidelity evaluation pipeline, where observed concept-response shifts are compared with surrogate-predicted shifts to assess local explanation fidelity.

The weighted error metrics are defined as:
\begin{equation}
\mathrm{WMSE}_i =
\frac{
\sum_{k=1}^{K} w^{(k)}
\left(\Delta y_i^{(k)} - \widehat{\Delta y}_i^{(k)}\right)^2
}{
\sum_{k=1}^{K} w^{(k)}
},
\label{eq:wmse}
\end{equation}

\begin{equation}
\mathrm{WMAE}_i =
\frac{
\sum_{k=1}^{K} w^{(k)}
\left|\Delta y_i^{(k)} - \widehat{\Delta y}_i^{(k)}\right|
}{
\sum_{k=1}^{K} w^{(k)}
}.
\label{eq:wmae}
\end{equation}

Here, $\widehat{\Delta y}_i^{(k)}$ denotes the surrogate-predicted concept-response shift for concept $c_i$ under perturbation $k$. The locality weight $w^{(k)}$ is computed using Eq.~\eqref{eq:locality_weight}. Therefore, the weighted error metrics in Eq.~\eqref{eq:wmse} and Eq.~\eqref{eq:wmae} give greater influence to perturbed samples that remain closer to the original input.

The weighted coefficient of determination is defined as:
\begin{equation}
R_{w,i}^{2}
=
1 -
\frac{
\sum_{k=1}^{K} w^{(k)}
\left(\Delta y_i^{(k)} - \widehat{\Delta y}_i^{(k)}\right)^2
}{
\sum_{k=1}^{K} w^{(k)}
\left(\Delta y_i^{(k)} - \overline{\Delta y}_{i,w}\right)^2
},
\label{eq:weighted_r2}
\end{equation}
where the weighted mean concept-response shift is given by:
\begin{equation}
\overline{\Delta y}_{i,w}
=
\frac{
\sum_{k=1}^{K} w^{(k)}\Delta y_i^{(k)}
}{
\sum_{k=1}^{K} w^{(k)}
}.
\label{eq:weighted_mean_shift}
\end{equation}

As defined in Eq.~\eqref{eq:weighted_r2}, $R_{w,i}^{2}$ measures how well the local surrogate explains the weighted variation in concept-response shifts. The weighted mean in Eq.~\eqref{eq:weighted_mean_shift} is used as the reference value for computing the weighted total variation.

\textbf{Faithfulness} measures whether the explanation is genuinely linked to the model's behaviour under perturbation. A faithful concept explanation should show larger response changes when concept-relevant regions are perturbed and smaller changes when irrelevant regions are modified. Faithfulness is measured using the Pearson correlation between concept-relevant perturbation strength and the absolute concept-response shift:
\begin{equation}
\mathrm{Faithfulness}_{i}
=
r\left(
\{s_i^{(k)}\}_{k=1}^{K},
\{|\Delta y_i^{(k)}|\}_{k=1}^{K}
\right),
\label{eq:faithfulness}
\end{equation}
where $s_i^{(k)}$ denotes the perturbation strength affecting concept $c_i$, and $r(\cdot,\cdot)$ denotes the Pearson correlation coefficient. As defined in Eq.~\eqref{eq:faithfulness}, higher correlation indicates that stronger perturbation of concept-relevant evidence produces larger concept-response changes.

\textbf{Stability} evaluates whether explanations remain similar when superficial non-clinical artefacts are added to the image. In this work, stability is tested using artefact-modified images, such as date-added and logo-added versions of the original input. The overlap between the original explanation set $A$ and the artefact-modified explanation set $B$ is measured using the Jaccard index:
\begin{equation}
J(A,B)=\frac{|A \cap B|}{|A \cup B|}.
\label{eq:jaccard_stability}
\end{equation}

As shown in Eq.~\eqref{eq:jaccard_stability}, higher Jaccard values indicate stronger stability because the explanation remains focused on the same concept-level evidence after non-clinical artefacts are introduced.

\textbf{Consistency} evaluates whether ConceptSMILE produces reproducible explanations when the same unchanged image is analysed repeatedly under identical experimental settings. Consistency is measured using the variance and standard deviation of concept-importance scores across repeated runs. Lower variance and lower standard deviation indicate that the explanation is more reproducible and less affected by random variation.

For repeated runs, the mean concept-importance score is first computed as:
\begin{equation}
\bar{a}_i =
\frac{1}{R}
\sum_{r=1}^{R} a_i^{(r)},
\label{eq:mean_importance}
\end{equation}
where $R$ denotes the number of repeated runs, $a_i^{(r)}$ denotes the concept-importance score for concept $c_i$ in run $r$, and $\bar{a}_i$ denotes the mean concept-importance score across repeated runs.

The consistency of the explanation is then quantified using the variance and standard deviation of the repeated concept-importance scores:
\begin{equation}
\mathrm{Var}_i =
\frac{1}{R-1}
\sum_{r=1}^{R}
\left(a_i^{(r)}-\bar{a}_i\right)^2,
\qquad
\mathrm{SD}_i=\sqrt{\mathrm{Var}_i}.
\label{eq:consistency}
\end{equation}

As defined in Eq.~\eqref{eq:mean_importance} and Eq.~\eqref{eq:consistency}, lower variance and standard deviation indicate stronger reproducibility across repeated executions.

Together, these five metrics provide a multi-dimensional audit of concept-level explanation reliability. Attribution accuracy tests whether the explanation points to meaningful evidence, fidelity evaluates whether the surrogate provides a reliable local approximation, faithfulness assesses whether importance reflects actual perturbation-response behaviour, stability tests robustness to irrelevant artefacts, and consistency measures reproducibility across repeated execution.

\subsection{Case Study and Experimental Protocol}
To demonstrate the proposed framework, ConceptSMILE is evaluated using retinal fundus image analysis as a high-stakes medical case study. The aim of this case study is not to train a new diagnostic classifier, but to examine whether concept-level explanations produced by existing model pathways remain reliable under controlled perturbation. Retinal imaging is selected because it contains clinically meaningful and visually grounded concepts, such as lesions, blood vessels, and the optic disc, which provide a suitable setting for evaluating concept-based explanations.

The experiments use retinal fundus images from four public datasets: HRF, APTOS, ODIR, and IDRiD. A controlled subset is selected from each dataset to support a balanced and comparable evaluation across different image sources, acquisition conditions, and retinal disease presentations. For each image, ConceptSMILE analyses three predefined retinal concepts: lesion, blood vessels, and optic disc.

Two concept-extraction pathways are evaluated under the same ConceptSMILE protocol. The MedSAM-based pathway provides segmentation-guided visual concept responses, while the vision-language-model pathway provides semantic concept responses using a fixed structured prompt. Both pathways are treated as fixed explanation systems. ConceptSMILE does not modify or retrain these models; instead, it audits their concept-level outputs through perturbation-based analysis.

For each image, the original concept responses are first recorded. The image is then divided into superpixels, and multiple perturbed versions are generated by masking selected regions. Each perturbed image is passed through the same concept-extraction pathway, and the resulting concept-response shifts are measured. Locality weights are then computed using embedding-space distances, and a local XGBoost surrogate is fitted to approximate the relationship between perturbation patterns and concept-response changes.

The reliability of each explanation is evaluated using attribution accuracy, fidelity, faithfulness, stability, and consistency. This experimental protocol allows ConceptSMILE to compare visual and semantic concept pathways under a unified auditing framework and to test whether concept-based explanations remain trustworthy when the input evidence is systematically changed.
\paragraph{Reproducibility details}
To support reproducibility, Table~\ref{tab:reproducibility_details} summarises the main experimental settings used in the ConceptSMILE evaluation, including the dataset subset, perturbation protocol, superpixel configuration, embedding model, distance metrics, surrogate model, random seeds, and VLM prompting setup. These settings describe the implementation used in this paper and are not intended to restrict ConceptSMILE to a specific model architecture or concept-extraction pathway.
\begin{table}[H]
\centering
\caption[Reproducibility details]{Reproducibility details for the ConceptSMILE experiments.}
\label{tab:reproducibility_details}

\begingroup
\smalltableformat

\begin{tabularx}{0.98\linewidth}{@{}p{0.25\linewidth}X@{}}
\toprule
\textbf{Component} & \textbf{Setting} \\
\midrule
Datasets & HRF, APTOS 2019, ODIR5K, and IDRiD \\
Evaluation subset & 40 retinal fundus images in total; 10 images selected from each dataset \\
Retinal concepts & Lesion, blood vessels, and optic disc \\
Perturbations & 50 unique binary superpixel perturbations per image; 2,000 perturbed samples in total; all-masked perturbation excluded \\
Superpixel settings & SLIC superpixels with target segments = 7 \\
Perturbation operation & Masked superpixels were removed by setting the corresponding image regions to black \\
Embedding model & \texttt{facebook/dinov2-base}; CLS-token embedding used for locality computation \\
Distance metrics & Cosine distance and Wasserstein distance \\
Locality weighting & Exponential locality kernel applied to convert distances into perturbation weights \\
VLM prompt format & Fixed structured prompt requiring a valid JSON response for the three concepts \\
Surrogate model & XGBoost regressor \\
\bottomrule
\end{tabularx}

\endgroup
\end{table}
\FloatBarrier
\section{Results}
This section presents the experimental evaluation of ConceptSMILE using retinal fundus image analysis as a high-stakes case study. The aim is not to assess diagnostic classification performance, but to examine whether concept-level explanations remain reliable when tested under controlled perturbation. Results are reported for two concept-extraction pathways: MedSAM-based visual concepts and vision-language-model-based semantic concepts. Both pathways are evaluated using the same ConceptSMILE protocol across three retinal concepts: lesion, blood vessels, and optic disc.The results are organised around five main reliability dimensions: attribution accuracy, fidelity, faithfulness, stability, and consistency, followed by an additional robustness analysis under acquisition artefacts.
\subsection{Qualitative Results}
\begin{figure}[H]
    \centering
    \includegraphics[width=0.8\linewidth]{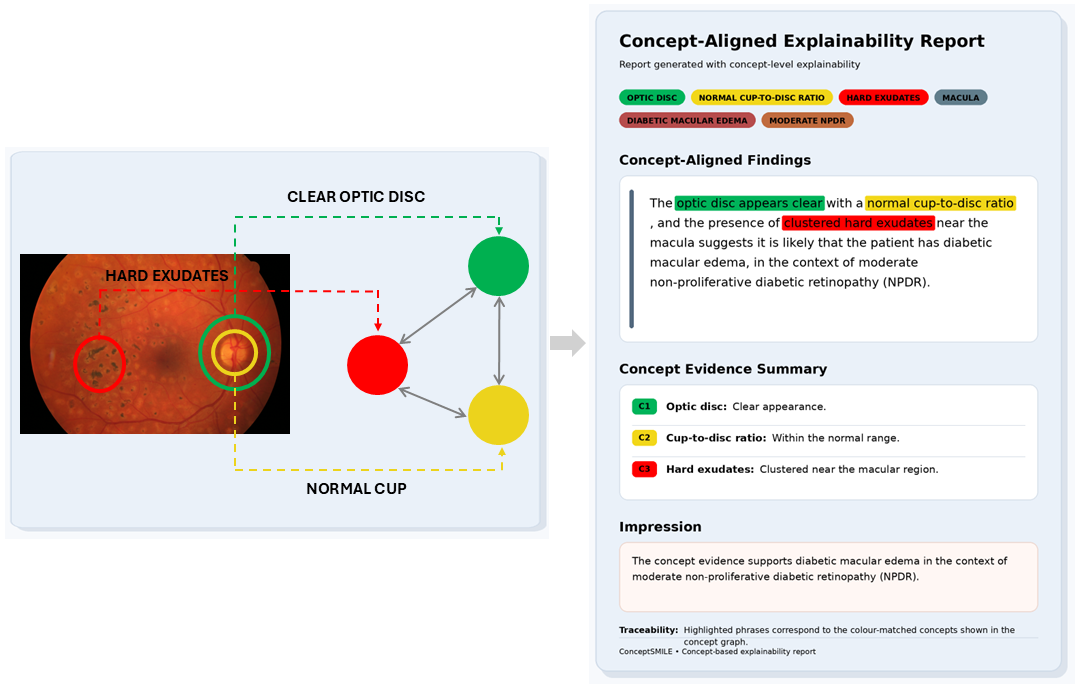}    
\caption{Qualitative ConceptSMILE output illustrating traceable alignment between retinal concept evidence, the concept graph, and the generated concept-level report. Colour-matched highlights link the clear optic disc, normal cup-to-disc ratio, and hard-exudate concepts to their corresponding textual evidence.}
    \label{fig:Qualitative ConceptSMILE output}
\end{figure}

Before reporting quantitative results, we first examine the type of concept-level explanations generated by ConceptSMILE. For each retinal image, the framework produces concept responses for lesion, blood vessels, and optic disc. In the MedSAM-based pathway, these concepts are represented through segmentation-guided visual regions and confidence scores. In the vision-language pathway, the same concepts are represented through structured semantic responses generated from the image.
Fig.~\ref{fig:Qualitative ConceptSMILE output} shows a representative qualitative ConceptSMILE output, illustrating how retinal concept evidence is linked to a concept graph and then to a generated concept-level report.
These qualitative outputs show that ConceptSMILE can represent retinal evidence at the concept level rather than only through low-level image regions or saliency maps. However, visual or semantic plausibility alone is not sufficient to establish explanation reliability. A concept may appear clinically meaningful while still being unstable, weakly faithful, or sensitive to irrelevant perturbations. Therefore, the following subsections evaluate whether these concept-level explanations remain accurate, faithful, stable, and consistent under systematic perturbation.
\subsection{Quantitative Results}
The quantitative evaluation assesses ConceptSMILE across five complementary reliability dimensions. Attribution accuracy examines whether important concept regions align with clinically meaningful evidence. Faithfulness evaluates whether concept responses change when relevant evidence is perturbed. Stability tests robustness to superficial non-clinical artefacts, while consistency measures reproducibility across repeated runs. Finally, attribution fidelity evaluates how well the local surrogate approximates concept-response behaviour under perturbation.
\subsubsection{Surrogate Model and Locality Selection}
Before reporting the main ConceptSMILE reliability results, we compared several local explanation variants to evaluate the effect of surrogate model choice and locality weighting. The evaluated variants included ConceptLIME, ConceptBayLIME, Wasserstein-weighted linear regression, cosine-weighted XGBoost, and Wasserstein-weighted XGBoost. The comparison was based on surrogate fidelity, using weighted MSE, weighted MAE, $R^2$, and weighted $R^2$.

Table~\ref{tab:surrogate_model_comparison} reports the averaged results across the repeated MedSAM and VLM experiments. For the MedSAM pathway, the Wasserstein-weighted XGBoost configuration achieved the strongest overall fidelity, with the lowest weighted error and the highest $R^2$ and weighted $R^2$. This supports the use of Wasserstein locality for segmentation-guided visual concepts, where distributional changes between original and perturbed retinal images are important. For the VLM pathway, the two XGBoost-based variants performed most strongly, but the advantage of Wasserstein locality was less uniform. ConceptSMILE achieved the highest ordinary $R^2$, whereas ConceptLIME-XGB achieved slightly lower weighted error and slightly higher weighted $R^2$. Therefore, the results support XGBoost as the final surrogate model, while also showing that locality choice remains pathway-dependent.

After this surrogate-selection stage, all subsequent fidelity results are reported using XGBoost with two locality definitions: cosine-weighted XGBoost and Wasserstein-weighted XGBoost. This design allows the effect of the distance function to be analysed while keeping the surrogate model fixed. In the remainder of the paper, cosine-weighted XGBoost is reported as the ConceptLIME-XGB baseline, while Wasserstein-weighted XGBoost is reported as the final ConceptSMILE configuration.
\begin{table}[H]
\centering
\caption[Surrogate model and locality-weighting comparison]{Averaged surrogate model and locality-weighting comparison for the MedSAM and VLM pathways.}
\label{tab:surrogate_model_comparison}

\begingroup
\resulttableformat

\begin{adjustbox}{max width=0.90\linewidth,center}
\begin{tabular}{@{}llccrrrr@{}}
\toprule
\textbf{Pathway} &
\textbf{Variant} &
\textbf{Dist.} &
\textbf{Surr.} &
\textbf{WMSE} &
\textbf{WMAE} &
\textbf{$R^2$} &
\textbf{$R_w^2$} \\
\midrule

\multirow{5}{*}{MedSAM}
& ConceptLIME      & C  & WLR & 0.0038 & 0.0255 & 0.5308 & 0.5662 \\
& ConceptBayLIME   & C  & BRR & 0.0037 & 0.0250 & 0.5868 & 0.6312 \\
& ConceptSMILE-WLR & WD & WLR & 0.0034 & 0.0263 & 0.6456 & 0.6561 \\
& ConceptLIME-XGB  & C  & XGB & \underline{0.0024} & \underline{0.0174} & \underline{0.6481} & \underline{0.7014} \\
& \textbf{ConceptSMILE} & WD & XGB & \textbf{0.0018} & \textbf{0.0146} & \textbf{0.8503} & \textbf{0.8465} \\
\midrule

\multirow{5}{*}{VLM}
& ConceptLIME      & C  & WLR & 0.0030 & 0.0373 & 0.3258 & 0.3481 \\
& ConceptBayLIME   & C  & BRR & 0.0028 & 0.0363 & 0.3839 & 0.3977 \\
& ConceptSMILE-WLR & WD & WLR & 0.0030 & 0.0379 & 0.3627 & 0.3597 \\
& \textbf{ConceptLIME-XGB} & C & XGB & \textbf{0.0022} & \textbf{0.0305} & \textbf{0.5127} & \textbf{0.5236} \\
& ConceptSMILE     & WD & XGB & \underline{0.0023} & \underline{0.0306} & \underline{0.5070} & \underline{0.5136} \\
\bottomrule
\end{tabular}
\end{adjustbox}

\vspace{0.5mm}

\begin{minipage}{0.90\linewidth}
\tiny
\textit{Note.} Lower WMSE and WMAE and higher $R^2$ and $R_w^2$ indicate stronger local surrogate fidelity. 
Best and second-best results within each pathway and metric are shown in bold and underlined, respectively. 
C = cosine distance; WD = Wasserstein distance; WLR = weighted linear regression; BRR = Bayesian ridge regression; XGB = XGBoost.
\end{minipage}

\endgroup
\end{table}
XGBoost was selected as the final surrogate model because concept-response behaviour under perturbation is unlikely to be purely linear. Masking one superpixel may affect multiple retinal concepts simultaneously, and lesion, vessel, or optic-disc responses may depend on interactions between neighbouring regions. While weighted linear regression and Bayesian ridge regression provide useful linear baselines, they may underfit these non-linear perturbation--response patterns. XGBoost can better capture feature interactions and sparse binary perturbation effects. Empirically, the XGBoost-based variants achieved stronger surrogate fidelity than the linear alternatives. Wasserstein-weighted XGBoost performed strongest for the MedSAM pathway, while the VLM pathway showed a smaller difference between cosine-weighted and Wasserstein-weighted XGBoost.

After surrogate selection, all subsequent fidelity results are reported using XGBoost with two locality definitions: cosine-weighted XGBoost and Wasserstein-weighted XGBoost. This allows the effect of locality weighting to be analysed while keeping the surrogate model fixed. Cosine-weighted XGBoost is reported as the ConceptLIME-XGB baseline, while Wasserstein-weighted XGBoost is reported as the final ConceptSMILE configuration. The results are therefore interpreted pathway-specifically rather than assuming that one locality measure is uniformly superior.
\subsubsection{Attribution Accuracy of Concept-Level Explanations}
Attribution accuracy evaluates whether the regions identified as important by ConceptSMILE correspond to clinically meaningful retinal evidence. In this study, attribution accuracy is assessed for three predefined concepts: lesion, blood vessels, and optic disc. The results are reported for both MedSAM-based visual concepts and vision-language-model-based semantic concepts using cosine and Wasserstein distance functions.
Table~\ref{tab:combined_att_accuracy} reports the concept-level attribution accuracy results for the MedSAM and VLM pathways across the four retinal datasets.
\begin{table}[H]
\centering
\caption[Concept-level attribution accuracy]{Concept-level attribution accuracy for the MedSAM and VLM pathways.}
\label{tab:combined_att_accuracy}

\begingroup
\resulttableformat

\begin{adjustbox}{max width=0.95\linewidth,center}
\begin{tabular}{@{}lllcccccc@{}}
\toprule
\textbf{Pathway} &
\textbf{Dataset} &
\textbf{Concept} &
\multicolumn{3}{c}{\textbf{Cosine}} &
\multicolumn{3}{c}{\textbf{Wasserstein}} \\
\cmidrule(lr){4-6}
\cmidrule(lr){7-9}
& & &
\textbf{ACC} &
\textbf{F1} &
\textbf{AUROC} &
\textbf{ACC} &
\textbf{F1} &
\textbf{AUROC} \\
\midrule

\multirow{12}{*}{\textbf{MedSAM}}

& \multirow{3}{*}{HRF}
& Lesion
& 0.4025 & 0.1203 & 0.2159
& 0.4096 & 0.1173 & 0.1820 \\

& & Vessel
& \underline{0.7394}
& \underline{0.7286}
& \underline{0.7131}
& \underline{0.7746}
& \textbf{0.7685}
& \textbf{0.7545} \\

& & Optic disc
& \textbf{0.8300}
& \textbf{0.8255}
& \textbf{0.7896}
& \textbf{0.8088}
& \underline{0.7475}
& \underline{0.6747} \\

\cmidrule(lr){2-9}

& \multirow{3}{*}{APTOS}
& Lesion
& \underline{0.6250}
& 0.4195
& 0.4056
& \underline{0.5175}
& 0.1910
& 0.2607 \\

& & Vessel
& \textbf{0.6927}
& \textbf{0.6777}
& \textbf{0.6214}
& \textbf{0.6154}
& \textbf{0.5740}
& \textbf{0.5850} \\

& & Optic disc
& 0.5400
& \underline{0.6230}
& \underline{0.5291}
& 0.4550
& \underline{0.4633}
& \underline{0.4122} \\

\cmidrule(lr){2-9}

& \multirow{3}{*}{ODIR5K}
& Lesion
& \textbf{0.7320}
& 0.4896
& 0.4137
& \underline{0.6462}
& 0.4630
& 0.3730 \\

& & Vessel
& \underline{0.7225}
& \underline{0.7056}
& \textbf{0.6537}
& \textbf{0.7205}
& \underline{0.7065}
& \textbf{0.6330} \\

& & Optic disc
& 0.6360
& \textbf{0.7423}
& \underline{0.6304}
& 0.6460
& \textbf{0.7100}
& \underline{0.6310} \\

\cmidrule(lr){2-9}

& \multirow{3}{*}{IDRiD}
& Lesion
& 0.4890
& 0.2139
& 0.2724
& \underline{0.5943}
& \underline{0.2524}
& \underline{0.3110} \\

& & Vessel
& \textbf{0.7781}
& \textbf{0.8050}
& \textbf{0.7555}
& \textbf{0.9240}
& \textbf{0.9116}
& \textbf{0.9080} \\

& & Optic disc
& \underline{0.5333}
& \underline{0.6411}
& \underline{0.4306}
& 0.4220
& 0.2010
& 0.2110 \\

\midrule

\multirow{12}{*}{\textbf{VLM}}

& \multirow{3}{*}{HRF}
& Lesion
& 0.5328
& 0.5414
& 0.0610
& \underline{0.4429}
& \underline{0.4437}
& \underline{0.1919} \\

& & Vessel
& \underline{0.5511}
& \underline{0.5512}
& \underline{0.0777}
& 0.0515
& 0.0478
& 0.0267 \\

& & Optic disc
& \textbf{0.7362}
& \textbf{0.7343}
& \textbf{0.4543}
& \textbf{0.7181}
& \textbf{0.7073}
& \textbf{0.2040} \\

\cmidrule(lr){2-9}

& \multirow{3}{*}{APTOS}
& Lesion
& 0.4762
& 0.4754
& \underline{0.4543}
& \underline{0.4791}
& \underline{0.4812}
& 0.0611 \\

& & Vessel
& \underline{0.5016}
& \underline{0.5453}
& \textbf{0.6632}
& 0.0583
& 0.0325
& \underline{0.1232} \\

& & Optic disc
& \textbf{0.6316}
& \textbf{0.6562}
& 0.1115
& \textbf{0.7383}
& \textbf{0.7313}
& \textbf{0.4501} \\

\cmidrule(lr){2-9}

& \multirow{3}{*}{ODIR5K}
& Lesion
& \underline{0.5740}
& 0.3663
& 0.5090
& \underline{0.5740}
& 0.3663
& 0.5090 \\

& & Vessel
& \textbf{0.6460}
& \textbf{0.4202}
& \textbf{0.5559}
& \textbf{0.6461}
& \textbf{0.4222}
& \textbf{0.5560} \\

& & Optic disc
& 0.5720
& \underline{0.4041}
& \underline{0.5230}
& 0.5720
& \underline{0.4041}
& \underline{0.5230} \\

\cmidrule(lr){2-9}

& \multirow{3}{*}{IDRiD}
& Lesion
& 0.5500
& \textbf{0.4376}
& \textbf{0.5162}
& 0.5860
& \textbf{0.4530}
& \textbf{0.5366} \\

& & Vessel
& \textbf{0.6100}
& \underline{0.4072}
& \underline{0.5141}
& \underline{0.6000}
& \underline{0.4154}
& \underline{0.5194} \\

& & Optic disc
& \underline{0.5840}
& 0.3846
& 0.4904
& \textbf{0.6020}
& 0.3784
& 0.4809 \\

\bottomrule
\end{tabular}
\end{adjustbox}

\vspace{0.5mm}

\begin{minipage}{0.95\linewidth}
\tiny
\textit{Note.} Best and second-best results within each pathway, dataset, distance metric, and evaluation metric are shown in bold and underlined, respectively. ACC = attribution accuracy; F1 = attribution F1 score; AUROC = attribution area under the receiver operating characteristic curve.
\end{minipage}

\endgroup
\end{table}
The MedSAM-based pathway generally produced stronger attribution accuracy for visually well-defined retinal structures. Blood vessels and optic disc regions achieved higher and more consistent attribution scores across several datasets, reflecting their clearer spatial structure in fundus images. Lesion attribution was more variable, which is expected because lesions are often small, sparse, and heterogeneous in appearance. This suggests that segmentation-guided concept extraction provides a stronger basis for spatially grounded explanation when the target concept has clear visual boundaries.

The vision-language pathway showed a different attribution pattern. Semantic concept explanations were able to identify clinically meaningful concepts, but their attribution accuracy was more variable across concepts and datasets. Optic disc explanations were relatively stronger in some datasets, while vessel and lesion explanations were less consistent under certain distance settings. This indicates that semantic concept responses can support concept-level interpretation, but they are not always spatially precise.

The attribution accuracy results show that concept-level explanations are pathway-dependent. MedSAM-based visual concepts provide stronger spatial alignment for structured retinal evidence, whereas VLM-based semantic concepts offer a more flexible but less spatially grounded explanation route. These findings support the central motivation of ConceptSMILE: concept-based explanations should not be accepted as automatically reliable simply because they are human-understandable, but should be evaluated according to how well they align with meaningful evidence.

\subsubsection{Faithfulness of Concept-Level Explanations}
Faithfulness evaluates whether the concept-level explanations produced by
ConceptSMILE are genuinely connected to the behaviour of the audited pathway.
A faithful explanation should show a clear relationship between perturbing
concept-relevant image regions and observing a corresponding change in the
concept response. In other words, if a concept is important to the explanation,
removing or masking evidence associated with that concept should produce a
measurable response shift.
Table~\ref{tab:combined_faithfulness} reports the faithfulness results for the MedSAM and VLM pathways under cosine and Wasserstein locality weighting.

\begin{table}[H]
\centering
\caption[Faithfulness of concept-level explanations]{Faithfulness of the MedSAM and VLM pathways under cosine and Wasserstein locality weighting.}
\label{tab:combined_faithfulness}

\begingroup
\resulttableformat

\begin{adjustbox}{max width=0.95\linewidth,center}
\begin{tabular}{@{}lllcccccc@{}}
\toprule
\textbf{Pathway} &
\textbf{Dataset} &
\textbf{Concept} &
\multicolumn{3}{c}{\textbf{Cosine}} &
\multicolumn{3}{c}{\textbf{Wasserstein}} \\
\cmidrule(lr){4-6}
\cmidrule(lr){7-9}
& & &
\textbf{$r$} &
\textbf{Std} &
\textbf{$p$} &
\textbf{$r$} &
\textbf{Std} &
\textbf{$p$} \\
\midrule

\multirow{12}{*}{\textbf{MedSAM}}

& \multirow{3}{*}{HRF}
& Lesion
& \textbf{0.6202} & \underline{0.2007} & \textbf{0.0041}
& \textbf{0.6554} & \underline{0.1845} & \textbf{0.0023} \\

& & Vessel
& \underline{0.4986} & 0.4196 & 0.0876
& \underline{0.5120} & 0.4020 & 0.0760 \\

& & Optic disc
& 0.3855 & \textbf{0.1075} & \underline{0.0241}
& 0.4238 & \textbf{0.0980} & \underline{0.0178} \\

\cmidrule(lr){2-9}

& \multirow{3}{*}{APTOS}
& Lesion
& \textbf{0.5401} & \underline{0.3630} & \textbf{0.0010}
& \textbf{0.5786} & \underline{0.3320} & \textbf{0.0006} \\

& & Vessel
& 0.1293 & 0.4470 & 0.2184
& 0.1418 & 0.4380 & 0.2030 \\

& & Optic disc
& \underline{0.4124} & \textbf{0.0000} & \underline{0.0864}
& \underline{0.4479} & \textbf{0.0000} & \underline{0.0690} \\

\cmidrule(lr){2-9}

& \multirow{3}{*}{ODIR5K}
& Lesion
& \textbf{0.4146} & \underline{0.3552} & \textbf{0.1182}
& \textbf{0.4472} & \underline{0.3340} & \textbf{0.0960} \\

& & Vessel
& \underline{0.3959} & 0.4432 & 0.1804
& \underline{0.4028} & 0.4310 & 0.1710 \\

& & Optic disc
& 0.3136 & \textbf{0.0000} & \underline{0.1239}
& 0.3524 & \textbf{0.0000} & \underline{0.0980} \\

\cmidrule(lr){2-9}

& \multirow{3}{*}{IDRiD}
& Lesion
& \underline{0.4908} & \underline{0.3859} & 0.0802
& \underline{0.5249} & \underline{0.3600} & 0.0620 \\

& & Vessel
& 0.4136 & 0.3956 & \underline{0.0625}
& 0.4245 & 0.3820 & \underline{0.0555} \\

& & Optic disc
& \textbf{0.4914} & \textbf{0.0000} & \textbf{0.0528}
& \textbf{0.5326} & \textbf{0.0000} & \textbf{0.0378} \\

\midrule

\multirow{12}{*}{\textbf{VLM}}

& \multirow{3}{*}{HRF}
& Lesion
& \underline{0.4455} & 0.2319 & \underline{0.1505}
& \underline{0.4318} & 0.2395 & \underline{0.1650} \\

& & Vessel
& \textbf{0.5978} & \textbf{0.1507} & \textbf{0.0067}
& \textbf{0.5902} & \textbf{0.1542} & \textbf{0.0082} \\

& & Optic disc
& 0.2219 & \underline{0.1736} & 0.2197
& 0.2136 & \underline{0.1810} & 0.2320 \\

\cmidrule(lr){2-9}

& \multirow{3}{*}{APTOS}
& Lesion
& 0.2241 & \underline{0.1691} & 0.2798
& 0.2185 & \underline{0.1726} & 0.2920 \\

& & Vessel
& \textbf{0.5794} & \textbf{0.1558} & \textbf{0.0149}
& \textbf{0.5711} & \textbf{0.1605} & \textbf{0.0172} \\

& & Optic disc
& \underline{0.3831} & 0.2272 & \underline{0.0741}
& \underline{0.3740} & 0.2350 & \underline{0.0818} \\

\cmidrule(lr){2-9}

& \multirow{3}{*}{ODIR5K}
& Lesion
& \underline{0.1820} & \underline{0.1584} & 0.3872
& \underline{0.1765} & \underline{0.1622} & 0.3990 \\

& & Vessel
& \textbf{0.5897} & \textbf{0.1171} & \textbf{0.0001}
& \textbf{0.5826} & \textbf{0.1224} & \textbf{0.0002} \\

& & Optic disc
& 0.1741 & 0.2312 & \underline{0.3548}
& 0.1669 & 0.2380 & \underline{0.3700} \\

\cmidrule(lr){2-9}

& \multirow{3}{*}{IDRiD}
& Lesion
& 0.2870 & 0.2796 & 0.2706
& 0.2762 & 0.2860 & 0.2830 \\

& & Vessel
& \textbf{0.7133} & \textbf{0.0632} & \textbf{0.0000}
& \textbf{0.7041} & \textbf{0.0680} & \textbf{0.0000} \\

& & Optic disc
& \underline{0.3228} & \underline{0.1652} & \underline{0.2114}
& \underline{0.3145} & \underline{0.1710} & \underline{0.2250} \\

\bottomrule
\end{tabular}
\end{adjustbox}

\vspace{0.5mm}

\begin{minipage}{0.95\linewidth}
\tiny
\textit{Note.} Higher Pearson $r$ indicates stronger faithfulness. Lower standard deviation and lower $p$-value indicate lower variability and stronger statistical evidence, respectively. Best and second-best results within each pathway and dataset are shown in bold and underlined.
\end{minipage}

\endgroup
\end{table}

The MedSAM pathway exhibited concept-dependent faithfulness rather than a
uniform advantage across all visually grounded structures. Lesion achieved the
highest Pearson correlation on HRF ($r=0.6202$), APTOS ($r=0.5401$), and ODIR
($r=0.4146$), while optic disc was marginally strongest on IDRiD
($r=0.4914$). However, statistically significant perturbation--response
relationships were observed only for HRF lesion ($p=0.0041$), HRF optic disc
($p=0.0241$), and APTOS lesion ($p=0.0010$). The correlations observed on ODIR
and IDRiD did not reach the conventional $p<0.05$ threshold, indicating that
moderate correlation values should not automatically be interpreted as strong
evidence of faithfulness.

The VLM pathway showed a clearer and more consistent concept-specific pattern.
The vessel concept achieved the highest Pearson correlation in all four
datasets, ranging from $r=0.5794$ on APTOS to $r=0.7133$ on IDRiD. These
relationships were statistically significant across HRF ($p=0.0067$), APTOS
($p=0.0149$), ODIR ($p=0.0001$), and IDRiD ($p=0.0001$). In contrast, lesion
and optic-disc concepts generally produced weaker correlations and
non-significant $p$-values. These findings show that semantic explanations were
not uniformly less faithful than segmentation-guided explanations; instead,
their reliability depended strongly on the concept being evaluated.

The faithfulness values under cosine and Wasserstein locality weighting were broadly similar, but not identical. This suggests that locality weighting had a limited effect on the perturbation--response correlations compared with the stronger influence of pathway, dataset, and concept type. The results demonstrate that faithfulness is both pathway- and concept-dependent. MedSAM provided stronger alignment for lesion evidence in several datasets, whereas the VLM pathway produced particularly consistent and statistically supported vessel explanations. These findings reinforce the central motivation of ConceptSMILE: concept-level explanations should be audited individually and should not be considered trustworthy solely because they are human-understandable.
\FloatBarrier
\subsubsection{Stability Under Non-Clinical Artefacts}
Stability evaluates whether concept-level explanations remain reliable when
superficial non-clinical artefacts are introduced into the image. In high-stakes
medical imaging, an explanation should remain focused on clinically meaningful
evidence rather than shifting in response to irrelevant visual elements such as
text, dates, logos, borders, or acquisition marks. In this experiment,
ConceptSMILE stability is assessed by comparing the explanation generated from
the original image with those generated after adding a date or an NHS-logo
artefact. The overlap is measured using the Jaccard indices $J_D$ and $J_L$,
where values closer to 1 indicate stronger stability.
Table~\ref{tab:stability_results} reports the stability results for the MedSAM and VLM pathways under date-added and NHS-logo artefact conditions.
The MedSAM pathway showed a clear concept-specific pattern. Optic-disc
explanations were the most stable MedSAM outputs across all four datasets under
the date-added condition, with $J_D$ values ranging from 0.96 to 0.98. The same
concept also achieved the highest MedSAM stability under the NHS-logo condition,
including 0.98 on HRF, 0.99 on APTOS, and 0.91 on ODIR. However, its stability
was lower on IDRiD under the logo condition ($J_L=0.45$). Vessel explanations
showed moderate stability under date artefacts, with values between 0.66 and
0.72, but were more sensitive to the logo artefact, where the corresponding
values ranged from 0.25 to 0.43. Lesion explanations were generally less stable
than optic-disc explanations, particularly on ODIR and IDRiD.

The VLM pathway exhibited a different stability profile. Lesion explanations
were highly stable under date-added artefacts on HRF, APTOS, ODIR, and IDRiD,
with $J_D$ values between 0.98 and 1.00. Vessel explanations were also highly
stable on ODIR and IDRiD, reaching 1.00 and 0.97, respectively. Under the
NHS-logo condition, VLM lesion stability reached 1.00 on APTOS and IDRiD, while
vessel stability reached 1.00 on ODIR and 0.99 on IDRiD. In contrast, VLM
optic-disc explanations were less robust, with values generally ranging from
0.43 to 0.67 across both artefact conditions.

These results demonstrate that stability cannot be assigned uniformly to an
entire explanation pathway. MedSAM provided particularly robust optic-disc
explanations, whereas the VLM pathway showed stronger stability for lesion and,
in several datasets, vessel concepts. Stability was therefore determined by the
interaction between the concept, pathway, dataset, and artefact type. Overall,
the findings support ConceptSMILE as an artefact-sensitivity audit that can
identify which concept-level explanations remain focused on retinal evidence
and which are more vulnerable to irrelevant visual modifications.
\begin{table}[H]
\centering
\caption[Stability under non-clinical artefacts]{Stability of the MedSAM and VLM pathways under date-added and NHS-logo artefacts.}
\label{tab:stability_results}

\begingroup
\resulttableformat

\begin{adjustbox}{max width=0.88\linewidth,center}
\begin{tabular}{@{}llcccc@{}}
\toprule
\textbf{Dataset} &
\textbf{Concept} &
\multicolumn{2}{c}{\textbf{Date artefact ($J_D$)}} &
\multicolumn{2}{c}{\textbf{NHS-logo artefact ($J_L$)}} \\
\cmidrule(lr){3-4}
\cmidrule(lr){5-6}
& &
\textbf{MedSAM} &
\textbf{VLM} &
\textbf{MedSAM} &
\textbf{VLM} \\
\midrule

\multirow{3}{*}{HRF}
& Lesion     & 0.60 & \textbf{0.99} & 0.49 & 0.67 \\
& Vessel     & 0.72 & 0.60 & 0.25 & \underline{0.68} \\
& Optic disc & \underline{0.97} & 0.66 & \textbf{0.98} & 0.43 \\

\cmidrule(lr){1-6}

\multirow{3}{*}{APTOS}
& Lesion     & 0.50 & \textbf{0.98} & 0.51 & \textbf{1.00} \\
& Vessel     & 0.66 & 0.66 & 0.43 & 0.43 \\
& Optic disc & \underline{0.96} & 0.43 & \underline{0.99} & 0.67 \\

\cmidrule(lr){1-6}

\multirow{3}{*}{ODIR5K}
& Lesion     & 0.43 & \textbf{1.00} & 0.43 & 0.67 \\
& Vessel     & 0.66 & \textbf{1.00} & 0.43 & \textbf{1.00} \\
& Optic disc & \underline{0.98} & 0.60 & \underline{0.91} & 0.43 \\

\cmidrule(lr){1-6}

\multirow{3}{*}{IDRiD}
& Lesion     & 0.43 & \textbf{0.98} & 0.25 & \textbf{1.00} \\
& Vessel     & 0.66 & \underline{0.97} & 0.43 & \underline{0.99} \\
& Optic disc & \textbf{0.98} & 0.67 & 0.45 & 0.43 \\

\bottomrule
\end{tabular}
\end{adjustbox}

\vspace{0.5mm}

\begin{minipage}{0.88\linewidth}
\tiny
\textit{Note.} Higher Jaccard values indicate stronger stability. Best and second-best results within each dataset and artefact condition are shown in bold and underlined, respectively. $J_D$ = Jaccard index under date-added artefact; $J_L$ = Jaccard index under NHS-logo artefact.
\end{minipage}

\endgroup
\end{table}
\subsubsection{Consistency Across Repeated Runs}
Consistency evaluates whether ConceptSMILE produces reproducible concept-level
explanations when the same unchanged image is analysed repeatedly under
identical experimental conditions. Unlike stability, which examines robustness
to non-clinical artefacts, consistency focuses on variation across repeated
executions of the original image. Lower variance and standard deviation indicate
that the concept-importance scores are less affected by stochastic perturbation
sampling or model-response variability.

In this study, consistency is evaluated by comparing the variation of
concept-importance scores across repeated runs. More reproducible explanations
show smaller changes in concept importance when the same image is processed
multiple times under the same settings.
Table~\ref{tab:combined_consistency} reports the repeated-run consistency results for the MedSAM and VLM pathways under cosine and Wasserstein locality weighting.
The MedSAM pathway generally produced low variation across repeated runs, but
the most consistent concept differed by dataset. On HRF and ODIR5K, optic disc
was the most reproducible concept under both cosine and Wasserstein weighting.
For HRF, optic-disc variance decreased from 0.000018 under cosine weighting to
0.000007 under Wasserstein weighting. On ODIR5K, optic disc also achieved the
lowest variance under both settings. In contrast, lesion was the most consistent
concept on APTOS, while the IDRiD result depended on the locality measure:
vessel was most consistent under cosine weighting, whereas lesion was most
consistent under Wasserstein weighting. These findings show that MedSAM
consistency was not limited to large anatomical structures, because lesion
explanations were also highly reproducible in some datasets.

The VLM pathway showed a similarly concept-dependent pattern. Optic disc
produced the lowest variance and standard deviation on HRF and APTOS under both
distance functions. However, lesion was the most consistent semantic concept on
ODIR5K and IDRiD. Vessel explanations generally showed greater variability than
the best-performing concept, while optic-disc explanations were particularly
variable on ODIR5K. Therefore, the generative VLM pathway did not exhibit
uniform instability; instead, reproducibility depended on the concept and
dataset being evaluated.

Comparing the two pathways, MedSAM achieved lower minimum variance values in
several experimental conditions, indicating stronger reproducibility for its
best-performing concepts. Nevertheless, neither pathway showed one concept that
was consistently dominant across all datasets. The choice of locality distance
also affected the ranking in some cases, particularly for MedSAM on IDRiD.

The consistency results demonstrate that reproducibility is both
pathway- and concept-dependent. MedSAM provided highly consistent explanations
for optic disc, lesion, or vessel concepts depending on the dataset, whereas the
VLM pathway was most reproducible for optic disc on HRF and APTOS and for
lesion on ODIR5K and IDRiD. These findings reinforce the need to evaluate
repeated explanation behaviour at the individual concept level rather than
assigning a single consistency judgement to an entire explanation pathway.
\begin{table}[H]
\centering
\caption[Consistency across repeated runs]{Consistency of the MedSAM and VLM pathways across repeated runs under cosine and Wasserstein locality weighting.}
\label{tab:combined_consistency}

\begingroup
\resulttableformat

\begin{adjustbox}{max width=0.96\linewidth,center}
\begin{tabular}{@{}lllcccc@{}}
\toprule
\textbf{Pathway} &
\textbf{Dataset} &
\textbf{Concept} &
\multicolumn{2}{c}{\textbf{Cosine}} &
\multicolumn{2}{c}{\textbf{Wasserstein}} \\
\cmidrule(lr){4-5}
\cmidrule(lr){6-7}
& & &
\textbf{Var.} $(\times 10^{-4})$ &
\textbf{Std} $(\times 10^{-3})$ &
\textbf{Var.} $(\times 10^{-4})$ &
\textbf{Std} $(\times 10^{-3})$ \\
\midrule

\multirow{12}{*}{\textbf{MedSAM}}

& \multirow{3}{*}{HRF}
& Lesion
& 0.8900 & 9.4090
& 0.8900 & 9.4460 \\

& & Vessel
& \underline{0.6300} & \underline{7.9270}
& \underline{0.2400} & \underline{4.8900} \\

& & Optic disc
& \textbf{0.1800} & \textbf{4.2420}
& \textbf{0.0700} & \textbf{2.6320} \\

\cmidrule(lr){2-7}

& \multirow{3}{*}{APTOS}
& Lesion
& \textbf{0.0100} & \textbf{0.0010}
& \textbf{0.0100} & \textbf{0.0010} \\

& & Vessel
& 1.9600 & 13.9900
& \underline{0.2400} & \underline{4.8800} \\

& & Optic disc
& \underline{0.2600} & \underline{5.0640}
& 0.5000 & 7.0590 \\

\cmidrule(lr){2-7}

& \multirow{3}{*}{ODIR5K}
& Lesion
& 1.1700 & 10.8220
& 0.6200 & 7.8770 \\

& & Vessel
& \underline{0.3300} & \underline{5.7480}
& \underline{0.3000} & \underline{5.5170} \\

& & Optic disc
& \textbf{0.0100} & \textbf{0.0110}
& \textbf{0.0100} & \textbf{0.0120} \\

\cmidrule(lr){2-7}

& \multirow{3}{*}{IDRiD}
& Lesion
& \underline{0.0900} & \underline{2.9410}
& \textbf{0.0300} & \textbf{1.8540} \\

& & Vessel
& \textbf{0.0800} & \textbf{2.7910}
& 0.9600 & 9.8000 \\

& & Optic disc
& 3.2100 & 17.9280
& \underline{0.0600} & \underline{2.5230} \\

\midrule

\multirow{12}{*}{\textbf{VLM}}

& \multirow{3}{*}{HRF}
& Lesion
& \underline{0.6100} & \underline{7.8320}
& \underline{0.3900} & \underline{6.2260} \\

& & Vessel
& 1.4400 & 11.9850
& 0.9400 & 9.6850 \\

& & Optic disc
& \textbf{0.4700} & \textbf{6.8820}
& \textbf{0.3400} & \textbf{5.8050} \\

\cmidrule(lr){2-7}

& \multirow{3}{*}{APTOS}
& Lesion
& \underline{0.4600} & \underline{6.7860}
& \underline{0.3800} & \underline{6.1770} \\

& & Vessel
& 0.7100 & 8.4120
& 0.6400 & 8.0100 \\

& & Optic disc
& \textbf{0.1600} & \textbf{4.0290}
& \textbf{0.1900} & \textbf{4.3190} \\

\cmidrule(lr){2-7}

& \multirow{3}{*}{ODIR5K}
& Lesion
& \textbf{0.4100} & \textbf{6.4030}
& \textbf{0.3600} & \textbf{5.9790} \\

& & Vessel
& \underline{1.2100} & \underline{10.9810}
& \underline{0.5700} & \underline{7.5580} \\

& & Optic disc
& 3.5900 & 18.9490
& 2.9800 & 17.2600 \\

\cmidrule(lr){2-7}

& \multirow{3}{*}{IDRiD}
& Lesion
& \textbf{0.3600} & \textbf{5.9960}
& \textbf{0.2800} & \textbf{5.2560} \\

& & Vessel
& 1.3800 & 11.7260
& \underline{0.5500} & \underline{7.3850} \\

& & Optic disc
& \underline{1.2600} & \underline{11.2270}
& 0.8400 & 9.1740 \\

\bottomrule
\end{tabular}
\end{adjustbox}

\vspace{0.5mm}

\begin{minipage}{0.96\linewidth}
\scriptsize
\textit{Note.} Variance and standard deviation are reported after scaling by $10^{-4}$ and $10^{-3}$, respectively, for readability. Lower variance and standard deviation indicate greater reproducibility. Best and second-best results within each pathway, dataset, and distance function are shown in bold and underlined, respectively. Var. = variance; Std = standard deviation.
\end{minipage}

\endgroup
\end{table}
\subsubsection{Attribution Fidelity}
\begin{figure}[H]
    \centering
    \includegraphics[width=0.85\linewidth]{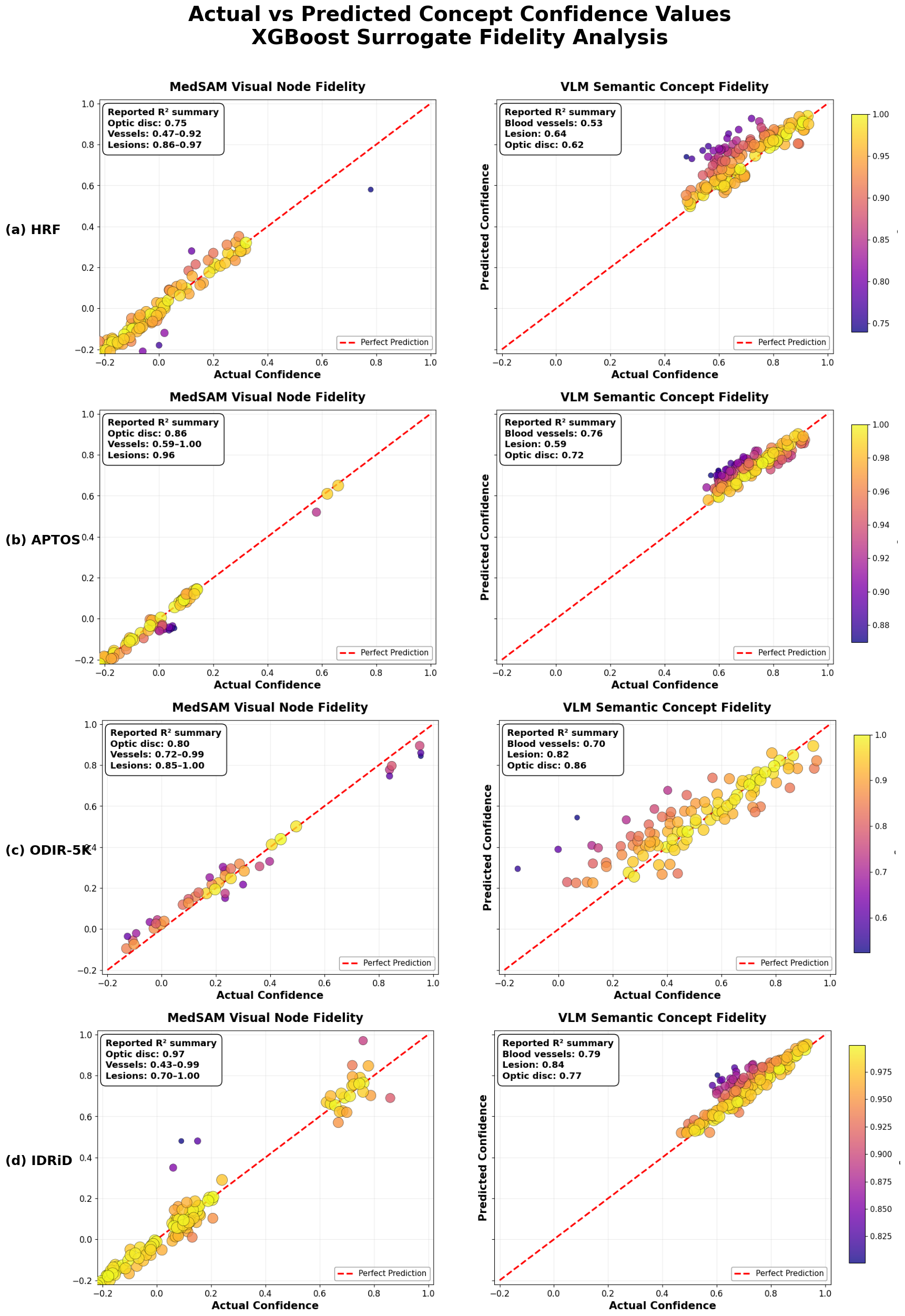}
    \caption{XGBoost surrogate fidelity comparison across HRF, APTOS, ODIR-5K, and IDRiD. The left column shows MedSAM-based visual node fidelity, and the right column shows VLM-based semantic concept fidelity. Points closer to the dashed diagonal indicate stronger agreement between actual and predicted concept-confidence values.}
    \label{fig:XGboost plot}
\end{figure}

Surrogate fidelity evaluates how well the local XGBoost surrogate approximates the concept-response behaviour observed under perturbation. In ConceptSMILE, high fidelity indicates that the surrogate can reproduce the relationship between superpixel perturbations and concept-confidence shifts. Therefore, fidelity provides evidence that the generated explanation is not only visually or semantically plausible, but also a reliable local approximation of the original concept pathway.
Fig.~\ref{fig:XGboost plot} presents a representative XGBoost surrogate-fidelity comparison between the MedSAM and VLM pathways.
As reported in Table~\ref{tab:combined_fidelity}, the MedSAM-based pathway generally produced stronger surrogate fidelity than the VLM pathway. This suggests that segmentation-guided visual concepts provide a more locally predictable response under superpixel perturbation. Concepts such as blood vessels and optic disc structures are spatially grounded in the fundus image, allowing the surrogate model to capture their perturbation-response behaviour more consistently. Lesion-related fidelity was more variable, which is expected because lesions are often smaller, sparser, and more heterogeneous across datasets.

\begin{table}[H]
\centering
\caption[Surrogate fidelity under cosine and Wasserstein weighting]{Surrogate fidelity for the MedSAM and VLM pathways under cosine and Wasserstein locality weighting.}
\label{tab:combined_fidelity}

\begingroup
\resulttableformat

\begin{adjustbox}{max width=\linewidth,center}
\begin{tabular}{@{}lllcccccccc@{}}
\toprule
\textbf{Pathway} &
\textbf{Dataset} &
\textbf{Concept} &
\multicolumn{4}{c}{\textbf{Cosine}} &
\multicolumn{4}{c}{\textbf{Wasserstein}} \\
\cmidrule(lr){4-7}
\cmidrule(lr){8-11}
& & &
\textbf{WMSE} &
\textbf{WMAE} &
\textbf{$R^2$} &
\textbf{$R_w^2$} &
\textbf{WMSE} &
\textbf{WMAE} &
\textbf{$R^2$} &
\textbf{$R_w^2$} \\
\midrule

\multirow{12}{*}{\textbf{MedSAM}}

& \multirow{3}{*}{HRF}
& Lesion
& 0.0046 & 0.0255 & \underline{0.7238} & \underline{0.7362}
& 0.0044 & 0.0218 & \underline{0.8936} & \underline{0.8819} \\

& & Vessel
& \underline{0.0012} & \underline{0.0158} & 0.6940 & 0.7246
& \textbf{0.0013} & \underline{0.0152} & 0.8324 & 0.8136 \\

& & Optic disc
& \textbf{0.0010} & \textbf{0.0075} & \textbf{0.8655} & \textbf{0.8629}
& \underline{0.0020} & \textbf{0.0098} & \textbf{0.9186} & \textbf{0.9173} \\

\cmidrule(lr){2-11}

& \multirow{3}{*}{APTOS}
& Lesion
& 0.0016 & 0.0168 & \textbf{0.8462} & \textbf{0.8614}
& 0.0013 & 0.0144 & \textbf{0.8920} & \textbf{0.9005} \\

& & Vessel
& \underline{0.0013} & \underline{0.0158} & 0.6305 & 0.6575
& \underline{0.0011} & \underline{0.0132} & 0.6790 & 0.6900 \\

& & Optic disc
& \textbf{0.0010} & \textbf{0.0107} & \underline{0.7765} & \underline{0.7872}
& \textbf{0.0007} & \textbf{0.0098} & \underline{0.8720} & \underline{0.8780} \\

\cmidrule(lr){2-11}

& \multirow{3}{*}{ODIR5K}
& Lesion
& 0.0021 & 0.0168 & \underline{0.8108} & \underline{0.8617}
& 0.0029 & 0.0172 & \underline{0.8795} & \underline{0.8952} \\

& & Vessel
& \underline{0.0014} & \underline{0.0156} & 0.7015 & 0.7715
& \underline{0.0013} & \underline{0.0148} & 0.8047 & 0.8168 \\

& & Optic disc
& \textbf{0.0009} & \textbf{0.0090} & \textbf{0.8200} & \textbf{0.8800}
& \textbf{0.0007} & \textbf{0.0088} & \textbf{0.9032} & \textbf{0.9124} \\

\cmidrule(lr){2-11}

& \multirow{3}{*}{IDRiD}
& Lesion
& \underline{0.0012} & \underline{0.0234} & \textbf{0.9270} & \textbf{0.9225}
& 0.0052 & 0.0291 & \textbf{0.8433} & \textbf{0.8700} \\

& & Vessel
& \textbf{0.0011} & \textbf{0.0175} & \underline{0.7101} & \underline{0.8429}
& \textbf{0.0015} & \textbf{0.0217} & \underline{0.7192} & \underline{0.7425} \\

& & Optic disc
& 0.0044 & 0.0476 & 0.5207 & 0.3346
& \underline{0.0017} & \underline{0.0250} & 0.6595 & 0.6810 \\

\midrule

\multirow{12}{*}{\textbf{VLM}}

& \multirow{3}{*}{HRF}
& Lesion
& \textbf{0.0019} & \underline{0.0347} & 0.5142 & 0.5077
& \textbf{0.0016} & \textbf{0.0308} & 0.4300 & 0.3898 \\

& & Vessel
& 0.0067 & 0.0498 & \underline{0.5584} & \underline{0.5604}
& 0.0030 & 0.0422 & \underline{0.5444} & \underline{0.6024} \\

& & Optic disc
& \underline{0.0019} & \textbf{0.0316} & \textbf{0.6408} & \textbf{0.6463}
& \underline{0.0020} & \underline{0.0321} & \textbf{0.6048} & \textbf{0.6223} \\

\cmidrule(lr){2-11}

& \multirow{3}{*}{APTOS}
& Lesion
& \underline{0.0027} & \underline{0.0288} & 0.5328 & 0.5414
& \underline{0.0016} & \underline{0.0308} & 0.4762 & 0.4754 \\

& & Vessel
& 0.0030 & 0.0389 & \underline{0.5511} & \underline{0.5512}
& 0.0026 & 0.0366 & \underline{0.5016} & \underline{0.5453} \\

& & Optic disc
& \textbf{0.0014} & \textbf{0.0265} & \textbf{0.7362} & \textbf{0.7343}
& \textbf{0.0013} & \textbf{0.0240} & \textbf{0.6316} & \textbf{0.6562} \\

\cmidrule(lr){2-11}

& \multirow{3}{*}{ODIR5K}
& Lesion
& \underline{0.0012} & \underline{0.0268} & \underline{0.4791} & \underline{0.4812}
& \underline{0.0012} & \underline{0.0271} & 0.4429 & \underline{0.4437} \\

& & Vessel
& 0.0066 & 0.0533 & 0.0583 & 0.0325
& 0.0066 & 0.0560 & \underline{0.5146} & 0.0478 \\

& & Optic disc
& \textbf{0.0009} & \textbf{0.0217} & \textbf{0.7383} & \textbf{0.7313}
& \textbf{0.0008} & \textbf{0.0193} & \textbf{0.7189} & \textbf{0.7073} \\

\cmidrule(lr){2-11}

& \multirow{3}{*}{IDRiD}
& Lesion
& \underline{0.0017} & \underline{0.0310} & \textbf{0.5993} & \textbf{0.6972}
& \textbf{0.0011} & \textbf{0.0261} & \underline{0.5376} & \underline{0.5716} \\

& & Vessel
& 0.0042 & 0.0402 & 0.1988 & 0.2342
& 0.0028 & 0.0382 & 0.3525 & 0.3907 \\

& & Optic disc
& \textbf{0.0013} & \textbf{0.0265} & \underline{0.5476} & \underline{0.5521}
& \underline{0.0017} & \underline{0.0304} & \textbf{0.6394} & \textbf{0.6349} \\

\bottomrule
\end{tabular}
\end{adjustbox}

\vspace{0.5mm}

\begin{minipage}{\linewidth}
\tiny
\textit{Note.} Lower WMSE and WMAE and higher $R^2$ and $R_w^2$ indicate stronger surrogate fidelity. Best and second-best results within each pathway, dataset, distance function, and metric are shown in bold and underlined, respectively. WMSE = weighted mean squared error; WMAE = weighted mean absolute error.
\end{minipage}

\endgroup
\end{table}

The vision-language pathway showed moderate but more variable fidelity. This variability reflects the different nature of semantic concept generation. Unlike MedSAM, which produces explicit visual masks, the VLM produces language-based concept responses that may not correspond directly to precise image regions. As a result, the local surrogate may find it harder to approximate semantic concept shifts from superpixel perturbation vectors alone. Nevertheless, the results show that ConceptSMILE can still be applied to semantic concept pathways and can reveal when their local explanation behaviour is less predictable.

The comparison between cosine and Wasserstein distance also shows that locality definition can influence surrogate fidelity. In several cases, Wasserstein distance improved the local approximation by better capturing distributional changes between original and perturbed image embeddings. However, the benefit was not uniform across all concepts or datasets, indicating that distance choice should be treated as part of the explanation audit rather than as a fixed assumption.
The fidelity results show that ConceptSMILE can quantify how reliably a local surrogate approximates concept-level explanation behaviour. The stronger fidelity of the MedSAM-based pathway suggests that visually grounded concepts are generally easier to audit through perturbation-based surrogate modelling, while the more variable VLM results highlight the need for explicit reliability testing of semantic explanations.
\subsection{Robustness Under Acquisition Artefacts}
To evaluate ConceptSMILE under imperfect retinal image acquisition, controlled perturbations were applied to a representative HRF fundus image. Two acquisition-related conditions were considered: contrast variation, representing changes in illumination or image quality, and simulated blink-related occlusion, representing partial loss of retinal visibility during image capture. Contrast factors of $0.6$, $0.8$, $1.0$, $1.2$, and $1.4$ were tested, where $1.0$ denotes the original image. Simulated occlusion levels of $0\%$, $1\%$, $2\%$, and $3\%$ of the retinal width were also applied. For each condition, ConceptSMILE was rerun using the same concept extraction, perturbation, locality-weighting, and XGBoost surrogate settings. The resulting Test $R^2$ values measure how well the local surrogate reproduces concept-response changes for blood vessels, lesions, and the optic disc. 
\begin{figure}[H]
    \centering
    \includegraphics[width=0.95\linewidth]{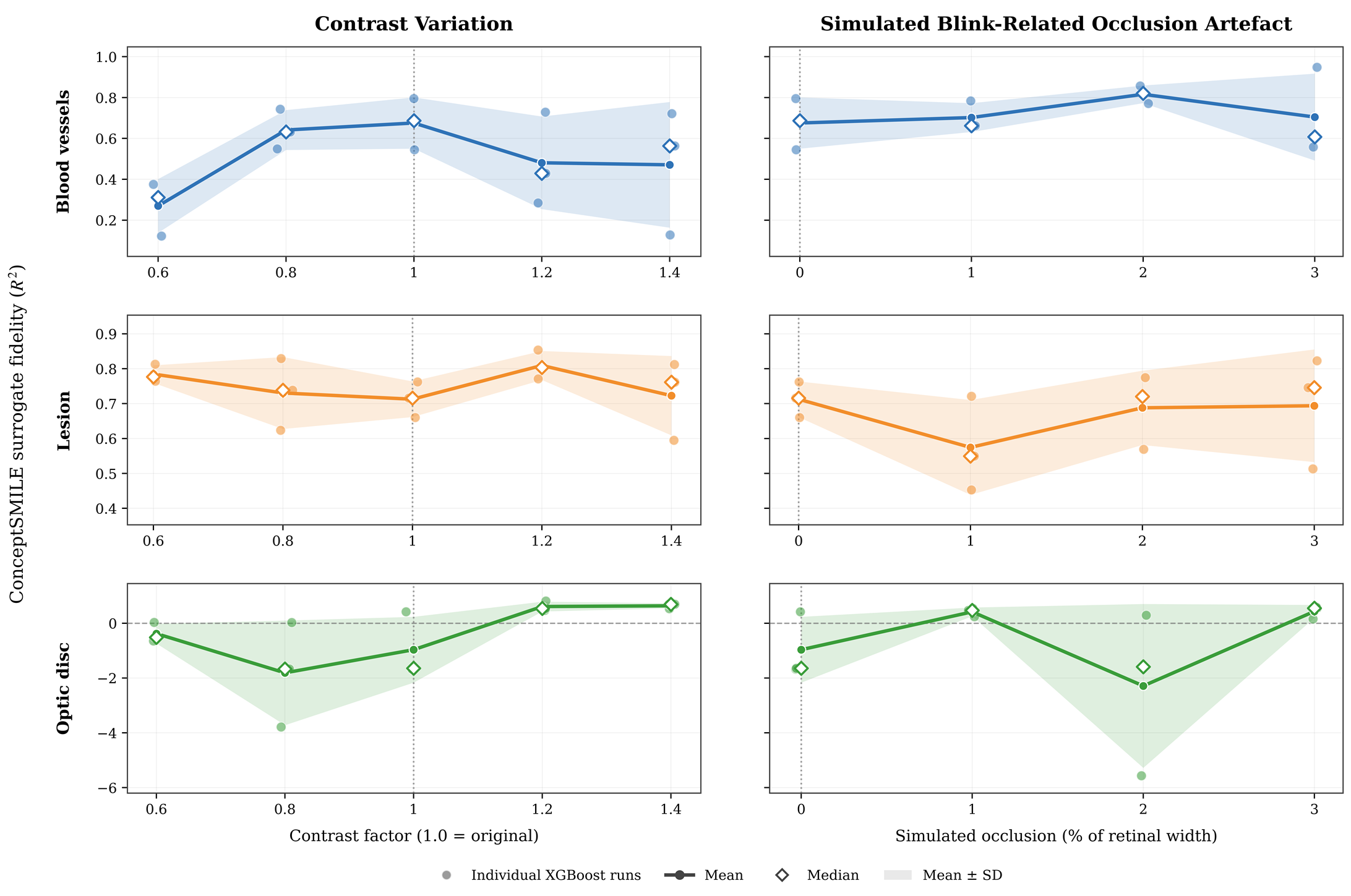}
    \caption{VLM-based ConceptSMILE robustness under contrast variation and simulated blink-related occlusion.}
\label{fig:conceptsmile_simulated_occlusion_robustness}
\end{figure}
\begin{figure}[H]
    \centering
    \includegraphics[width=0.95\linewidth]{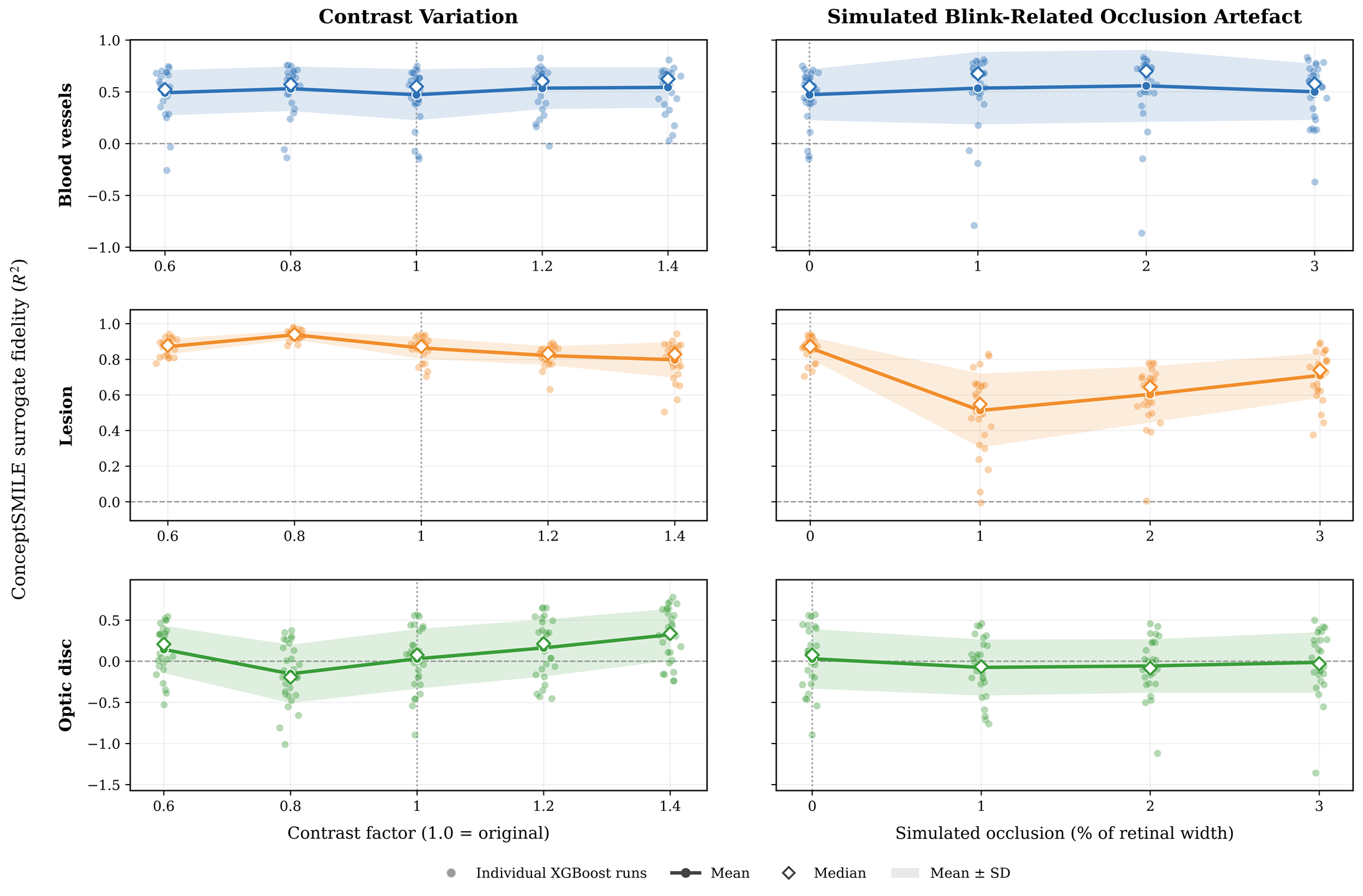}
    \caption{MedSAM-based ConceptSMILE robustness under contrast variation and simulated blink-related occlusion.}
\label{fig:medsam_corrected_acquisition_robustness}
\end{figure}
In Figures~\ref{fig:conceptsmile_simulated_occlusion_robustness} and~\ref{fig:medsam_corrected_acquisition_robustness}, individual points show repeated XGBoost train--test evaluations, solid lines show the mean, diamond markers indicate the median, and shaded regions represent mean $\pm$ standard deviation. The results show that acquisition artefacts affect retinal concepts differently. Blood-vessel and lesion explanations generally retain positive surrogate fidelity across most contrast and occlusion conditions, indicating that their local explanation behaviour remains predictable under moderate image-quality variation. In contrast, the optic-disc concept shows greater variability and occasional negative Test $R^2$ values, meaning that the surrogate performs worse than a mean-response baseline in those cases. This demonstrates that ConceptSMILE can act as a reliability-auditing mechanism by identifying which clinical concepts remain robust and which become unreliable when retinal images are affected by low contrast, partial occlusion, patient movement, or blink-related acquisition artefacts.
\subsection{Overall Findings}
The combined results demonstrate that the reliability of concept-based
explanations is not determined solely by whether the explanation is
human-understandable. Instead, reliability varies across concepts, model
pathways, datasets, locality measures, and perturbation conditions. This
supports the central premise of ConceptSMILE: concept explanations should be
treated as testable claims and evaluated across multiple complementary
dimensions.

The MedSAM pathway generally achieved stronger attribution accuracy and
surrogate fidelity for spatially defined retinal structures, particularly
blood vessels and the optic disc. However, its advantage was not uniform.
Faithfulness was strongest for lesion evidence in several datasets, while
stability and consistency depended on the concept and experimental condition.
MedSAM optic-disc explanations were particularly robust to non-clinical
artefacts and highly reproducible on HRF and ODIR, whereas lesion or vessel
concepts were more consistent in other datasets.

The VLM pathway showed a complementary reliability profile. Although its
spatial attribution and surrogate fidelity were generally more variable, its
vessel explanations achieved the strongest and most consistently significant
faithfulness correlations across all four datasets. The VLM also produced
highly stable lesion and vessel explanations under several date- and
logo-added conditions. Its consistency was strongest for optic disc on HRF and
APTOS and for lesion on ODIR and IDRiD. These findings show that semantic
concept explanations are not inherently less reliable, but their strengths
occur in different reliability dimensions from segmentation-guided concepts.

Cosine and Wasserstein locality weighting produced different fidelity and
consistency outcomes in several cases, although the faithfulness values were
unchanged. Distance selection should therefore be regarded as part of the audit
design rather than a fixed implementation choice. Overall, ConceptSMILE
revealed that no single pathway or concept was consistently superior across all
metrics. Trustworthy concept-based explanation consequently requires
multi-dimensional, concept-specific evaluation rather than reliance on visual
plausibility or a single aggregate score.
\section{Limitations and Future Mitigation}
Although ConceptSMILE provides a structured framework for auditing concept-based explanations, several limitations should be acknowledged. First, the empirical evaluation was limited in scale. The study used four publicly available diabetic-retinopathy-related retinal fundus datasets, namely HRF, APTOS, ODIR, and IDRiD, with 10 images selected from each dataset. Although this provides diversity across image sources, it remains a controlled proof-of-concept evaluation rather than a large-scale clinical validation. Future work should evaluate ConceptSMILE on larger multi-centre datasets, include a wider range of disease severity levels, and test whether the observed reliability patterns remain consistent across different populations, imaging devices, and clinical settings.

Second, the current study focused on three retinal concepts: lesion, blood vessels, and optic disc. These concepts are suitable for demonstrating the framework, but they do not capture the full complexity of retinal diagnosis. Important biomarkers such as microaneurysms, haemorrhages, hard exudates, macular abnormalities, and cup-to-disc changes were not evaluated as separate concepts. Future work should extend the concept vocabulary and assess whether ConceptSMILE remains reliable for small, sparse, overlapping, or spatially fragmented retinal findings.

Third, the framework depends on the quality of the underlying concept-extraction pathways. In this study, MedSAM was used for segmentation-guided visual concepts, while the VLM pathway was used for semantic concept responses. If either pathway produces incomplete, uncertain, or prompt-sensitive outputs, ConceptSMILE audits those outputs but does not automatically correct them. Future work should compare multiple segmentation models and vision--language models, incorporate uncertainty estimation, and explore ensemble-based concept extraction.

A related practical limitation concerns access to commercial vision--language models. During preliminary experiments, Gemini was explored as an additional semantic explanation pathway. However, the repeated perturbation-based querying required by ConceptSMILE appeared to trigger provider-side safety or usage restrictions, because the large number of modified retinal images could be interpreted as abnormal or attack-like automated behaviour. As a result, Gemini could not be included in the full experimental evaluation. Future work should investigate robust and compliant evaluation protocols for commercial multimodal models, including approved batch-testing access, clearer audit-use declarations, and reproducible logging of API restrictions.

Fourth, the perturbation strategy may not always produce clinically realistic image changes. Superpixel masking provides a practical way to test local concept-response behaviour, but it can introduce artificial patterns that may not correspond to plausible retinal counterfactuals. This is particularly important for small lesions, diffuse abnormalities, or structures whose interpretation depends on surrounding context. Future work should investigate anatomy-aware perturbations, lesion-aware masking, inpainting-based perturbations, and generative counterfactual methods that better preserve retinal realism.

Fifth, ConceptSMILE has a non-trivial computational cost. Each image requires superpixel generation, repeated perturbation, concept extraction, embedding computation, distance calculation, locality weighting, and surrogate fitting. This cost increases further when repeated runs are used for consistency analysis or when both MedSAM and VLM pathways are evaluated. Future implementations should improve efficiency through batching, caching, adaptive perturbation sampling, reduced surrogate-search spaces, and hardware-aware optimisation.

Sixth, the current evaluation does not provide causal proof. High surrogate fidelity or strong faithfulness indicates that concept responses are locally associated with perturbation behaviour, but it does not prove that the concept is causally responsible for the model output. Future work should combine ConceptSMILE with causal concept intervention, counterfactual testing, and clinician-verified concept manipulation to better distinguish correlation from causation.

Finally, this study did not include clinician-in-the-loop validation or prospective clinical testing. The reported metrics evaluate technical reliability dimensions such as attribution accuracy, fidelity, faithfulness, stability, and consistency, but they do not establish whether the explanations are useful, understandable, or actionable for ophthalmologists in real clinical workflows. Future work should include expert evaluation, reader studies, usability assessment, and prospective testing to determine whether ConceptSMILE explanations improve error detection, clinical trust calibration, and appropriate reliance on AI-assisted retinal interpretation.

The current study should be interpreted as an initial technical demonstration of concept-level reliability auditing. Further validation is needed to establish broader clinical generalisation, richer concept coverage, improved perturbation realism, computational scalability, causal evidence, and clinician-centred usefulness.

\section{Conclusion}

This paper introduced ConceptSMILE, a perturbation-based framework for auditing the reliability of concept-based explanations. Its central premise is that human-understandable concepts are not, by themselves, evidence of trustworthy model behaviour. A concept may appear clinically or semantically meaningful, but still be spatially inaccurate, weakly faithful to perturbation behaviour, sensitive to irrelevant artefacts, or inconsistent across repeated runs. ConceptSMILE addresses this problem by treating concept explanations as testable claims and evaluating them across attribution accuracy, surrogate fidelity, faithfulness, stability, and consistency.
The framework extends the SMILE-style local explanation principle from feature- or region-level attribution to concept-level reliability auditing. Rather than training a new concept model, ConceptSMILE audits existing concept-extraction pathways. It measures how concept responses change under controlled perturbations and uses a local surrogate model to approximate perturbation behaviour. This provides a common evaluation protocol for comparing visually grounded and semantically generated concept explanations.
The retinal case study showed that explanation reliability varied substantially across pathways, concepts, datasets, and evaluation metrics. MedSAM tended to provide stronger visually grounded attribution and local surrogate fidelity, particularly for vessel and optic-disc concepts. This suggests that segmentation-based visual concepts can provide reliable local evidence when the concept is spatially well defined. However, this advantage did not extend uniformly across all reliability dimensions.

The VLM pathway showed a complementary reliability profile. Although its spatial attribution and surrogate fidelity were generally weaker, it showed stronger faithfulness patterns for vessel-related semantic concepts and high stability for several lesion and vessel explanations under non-clinical artefacts. These results suggest that semantic concept responses can remain perturbation-sensitive and robust in some settings, even when they are less spatially precise than segmentation-based explanations.
The consistency analysis further showed that reproducibility should not be judged at the pathway level alone. The most reproducible concept changed across datasets and locality-weighting settings, with optic disc, lesion, and vessel concepts each showing stronger consistency under different conditions. These findings support concept-level reliability assessment rather than a single global trustworthiness score for an explanation pathway.

The results demonstrate that visual and semantic concept explanations have different reliability profiles. An explanation may be stable but not spatially accurate, faithful but not highly reproducible, or locally well approximated by a surrogate while remaining weak in another reliability dimension. The main contribution of ConceptSMILE is to make these metric-specific strengths and weaknesses visible and measurable through a structured audit process.
ConceptSMILE does not provide clinical certification or causal verification. Instead, it provides quantitative audit evidence that can help identify fragile, poorly grounded, unstable, or inconsistent concept explanations before they are relied upon in high-stakes settings. 

In practice, this type of audit can help researchers, clinical experts, and AI developers move from
apparent interpretability toward better-grounded confidence in concept explanations. From a Responsible AI perspective, these findings show that interpretability should not be treated as equivalent to trustworthiness. ConceptSMILE provides a practical audit layer for checking whether concept-based explanations remain reliable under perturbation, artefact exposure, and repeated execution. This is particularly important in high-stakes domains, where apparently plausible explanations may otherwise encourage inappropriate trust in unreliable model behaviour. Future work should evaluate the
framework on larger and more diverse datasets, extend the concept vocabulary, incorporate clinician-validated reference
annotations, compare alternative perturbation and masking strategies, and test concept-level auditing in additional
medical and safety-critical applications.
\section*{List of Abbreviations}
\begin{center}
\small
\setlength{\tabcolsep}{8pt}
\renewcommand{\arraystretch}{1.10}

\begin{tabular}{p{0.22\textwidth} p{0.65\textwidth}}
\hline
\textbf{Abbreviation} & \textbf{Definition} \\
\hline
XAI & Explainable Artificial Intelligence \\
C-XAI & Concept-Based Explainable Artificial Intelligence \\
SMILE & Statistical Model-Agnostic Interpretability with Local Explanations \\
ConceptSMILE & Concept-level extension of SMILE proposed in this paper \\
MedSAM & Medical Segment Anything Model \\
VLM & Vision--Language Model \\
Qwen2.5-VL & Qwen2.5 Vision--Language Model \\
KG & Knowledge Graph \\
LIME & Local Interpretable Model-Agnostic Explanations \\
WD & Wasserstein Distance \\
WLR & Weighted Linear Regression \\
BRR & Bayesian Ridge Regression \\
XGB & XGBoost \\
ATT ACC & Attribution Accuracy \\
ATT F1 & Attribution F1 Score \\
ATT AUROC & Attribution Area Under the Receiver Operating Characteristic Curve \\
MSE & Mean Squared Error \\
MAE & Mean Absolute Error \\
WMSE & Weighted Mean Squared Error \\
WMAE & Weighted Mean Absolute Error \\
$R^2$ & Coefficient of Determination \\
$R_w^2$ & Weighted Coefficient of Determination \\
$J_D$ & Jaccard index under date-added artefact condition \\
$J_L$ & Jaccard index under NHS-logo artefact condition \\
\hline
\end{tabular}
\end{center}
\clearpage
\bibliographystyle{unsrt}  
\bibliography{references}
\end{document}